\documentclass{article}

\usepackage{arxiv}

\usepackage[utf8]{inputenc} 
\usepackage[T1]{fontenc}    
\usepackage[hypertexnames=true]{hyperref}       
\usepackage{url}            
\usepackage{booktabs}       
\usepackage{amsmath, amsfonts}       
\usepackage{amsthm}
\usepackage{nicefrac}       
\usepackage{microtype}      
\usepackage{cleveref}       
\usepackage{graphicx}
\usepackage{doi}

\usepackage{amssymb}
\usepackage{algorithmic}
\usepackage[ruled,linesnumbered,vlined]{algorithm2e}
\usepackage{array}
\usepackage[caption=false,font=normalsize,labelfont=sf,textfont=sf]{subfig}
\usepackage{textcomp}
\usepackage{stfloats}
\usepackage{verbatim}
\usepackage{cite}
\usepackage{multirow} 
\usepackage{scalerel}
\usepackage{tikz}
\usepackage[title,titletoc]{appendix}
\usepackage{multirow}
\usepackage{amssymb}
\usepackage{pifont}
\usepackage{enumitem}

\DeclareMathOperator*{\argmin}{\arg\min}

\theoremstyle{plain}

\theoremstyle{definition}

\theoremstyle{remark}

\title{Optimization-Driven Design of \\ Monolithic Soft-Rigid Grippers}

\author{
	\hspace{1mm}Pierluigi Mansueto \\
	Department of Information Engineering \\
	University of Florence \\
	Firenze, 50139, Italy \\
	\texttt{pierluigi.mansueto@unifi.it} \\
	\And
	\hspace{1mm}Mihai Dragusanu \\
	Department of Information Engineering \\
	and Mathematics, University of Siena \\ 
	Siena, 53100, Italy \\
	\texttt{mihai.dragusanu2@unisi.it} \\
	\And
	\hspace{1mm}Anjum Saeed \\
	Department of Information Engineering \\
	and Mathematics, University of Siena \\ 
	Siena, 53100, Italy \\
	\texttt{a.saeed1@student.unisi.it} \\
	\And
	\hspace{1mm}Monica Malvezzi\thanks{Humanoids \& Human Centered Mechatronics, Istituto Italiano di Tecnologia, Genoa, 16163, Italy} \\
	Department of Information Engineering \\
	and Mathematics, University of Siena \\ 
	Siena, 53100, Italy \\
	\texttt{monica.malvezzi@unisi.it} \\
	\And
	\hspace{1mm}Matteo Lapucci \\
	Department of Information Engineering \\
	University of Florence \\
	Firenze, 50139, Italy \\
	\texttt{matteo.lapucci@unifi.it} \\
	\And
	\hspace{1mm}Gionata Salvietti$^\ast$ \\
	Department of Information Engineering \\
	and Mathematics, University of Siena \\ 
	Siena, 53100, Italy \\
	\texttt{gionata.salvietti@unisi.it} \\
}

\renewcommand{\headeright}{Mansueto et al.}
\renewcommand{\undertitle}{}
\renewcommand{\shorttitle}{Optimization-Driven Design of Soft-Rigid Grippers}

\hypersetup{
pdftitle={Optimization-Driven Design of Monolithic Soft-Rigid Grippers},
pdfsubject={},
pdfauthor={Pierluigi Mansueto, Mihai Dragusanu, Anjum Saeed, Monica Malvezzi, Matteo Lapucci, Gionata Salvietti},
pdfkeywords={Soft-rigid grippers, sim-to-real design of soft grippers, compliant mechanisms},
}

\begin{document}
\maketitle

\begin{abstract}
	Sim-to-real transfer remains a significant challenge in soft robotics due to the unpredictability introduced by common manufacturing processes such as 3D printing and molding. These processes often result in deviations from simulated designs, requiring multiple prototypes before achieving a functional system. 
	In this study, we propose a novel methodology to address these limitations by combining advanced rapid prototyping techniques and an efficient optimization strategy. Firstly, we employ rapid prototyping methods typically used for rigid structures, leveraging their precision to fabricate compliant components with reduced manufacturing errors. Secondly, our optimization framework minimizes the need for extensive prototyping, significantly reducing the iterative design process. The methodology enables the identification of stiffness parameters that are more practical and achievable within current manufacturing capabilities. The proposed approach demonstrates a substantial improvement in the efficiency of prototype development while maintaining the desired performance characteristics. This work represents a step forward in bridging the sim-to-real gap in soft robotics, paving the way towards a faster and more reliable deployment of soft robotic systems.
\end{abstract}

\keywords{Soft-rigid grippers \and sim-to-real design of soft grippers \and compliant mechanisms}

\section{Introduction}
Soft multi-fingers hands~\cite{ deimel2013compliant,catalano2014adaptive,feng2018soft} and grippers~\cite{shintake2018soft} have demonstrated to be an efficient solution to the usability problem of the first generation of rigid robotic hands and manipulation mechanisms~\cite{Bicchi2000TR0, Prattichizzo08}. The use of flexible and deformable components allow robotic hands to adapt more naturally and effectively to the surrounding environment also addressing challenges arising from sensor uncertainties, particularly in defining the position and shape of the object. In fact, differently than traditional approaches that require precise contact points locations and force application, soft hands exploit their flexibility to passively adapt to various objects shape~\cite{PoMaSaBiMaPr_IJRR2020}.\\
Using soft materials offer numerous advantages, including improved safety in interactions with humans and reduced damage occurred to fragile objects during manipulation tasks.
The compliance in these robotic hands can be achieved through various techniques, such as advanced 3D-printing techniques~\cite{hussain2020design} and the use of deformable bio-materials~\cite{jones2021bubble}.
However, these fabrication techniques are prone to possible precision errors often leading to prototypes that presents mechanical characteristics different from those designed. Moreover, in more complex manipulation challenges, soft hands may face limitations in terms of dexterity~\cite{della2018toward}. These limitations arise from underactuation, which implies a reduced ability to apply high forces, and the difficulty in performing precise contact modeling~\cite{PoSaBiMaPr-ral2018}.

\textit{Soft-rigid} grippers have been recently introduced to try to partially solve the above mentioned issues~\cite{odhner2014compliant, dollar2010highly, Salvietti-Robosoft2020}. These end-effectors are based on the combination of rigid structures and soft joints to grip and manipulate objects. The underlying principle for the actuation system is to use a tendon transmission system, where the gripper fingers are defined by the concatenation of modules made of rigid material (ABS, PLA) and soft joints~\cite{malvezzi2019design}. This configuration consent to achieve different flexion trajectories of the fingers by varying the stiffness of the soft joints~\cite{SaHuMaPr-RAL2017} increasing the dexterity of the grippers. Stiffness of soft joints can be selected in different ways, varying, for instance the geometry or the material used (silicons with different hardness). An interesting opportunity come from 3D printing techniques, where it is possible to act on the \textit{infill density} during the printing process.
Commercial 3D printer software generally offers the ability to select infill density as a parameter during the additive manufacturing process. However, the accuracy and repeatability of stiffness values for soft joints is not very high, unless still expensive multi-material 3D printers are used~\cite{HuIqMaPrSa_HFR19}.

This limitation underline the importance of advanced design tools, which
enable the synthesis of mechanisms using a minimal set of mechanical components, where each component's shape is carefully tailored to ensure the intended mobility functions~\cite{xu2018bend,megaro2017computational}. Recently, these tools have been leveraged to design compliant robotic grippers and hands~\cite{mutlu20173d,8868668}. Mutlu et al.~\cite{mutlu2015effect} modeled and experimentally evaluated the bending behavior of established flexure hinge designs to determine the optimal configuration for a fully compliant prosthetic finger. Liu et al.~\cite{liu2018soft} presented a two-fingered gripper for fruit handling, where each finger features a monolithic, compliant structure shaped through topology optimization.
\begin{figure}[tbp]
	\centering
	\includegraphics[width=0.7\linewidth]{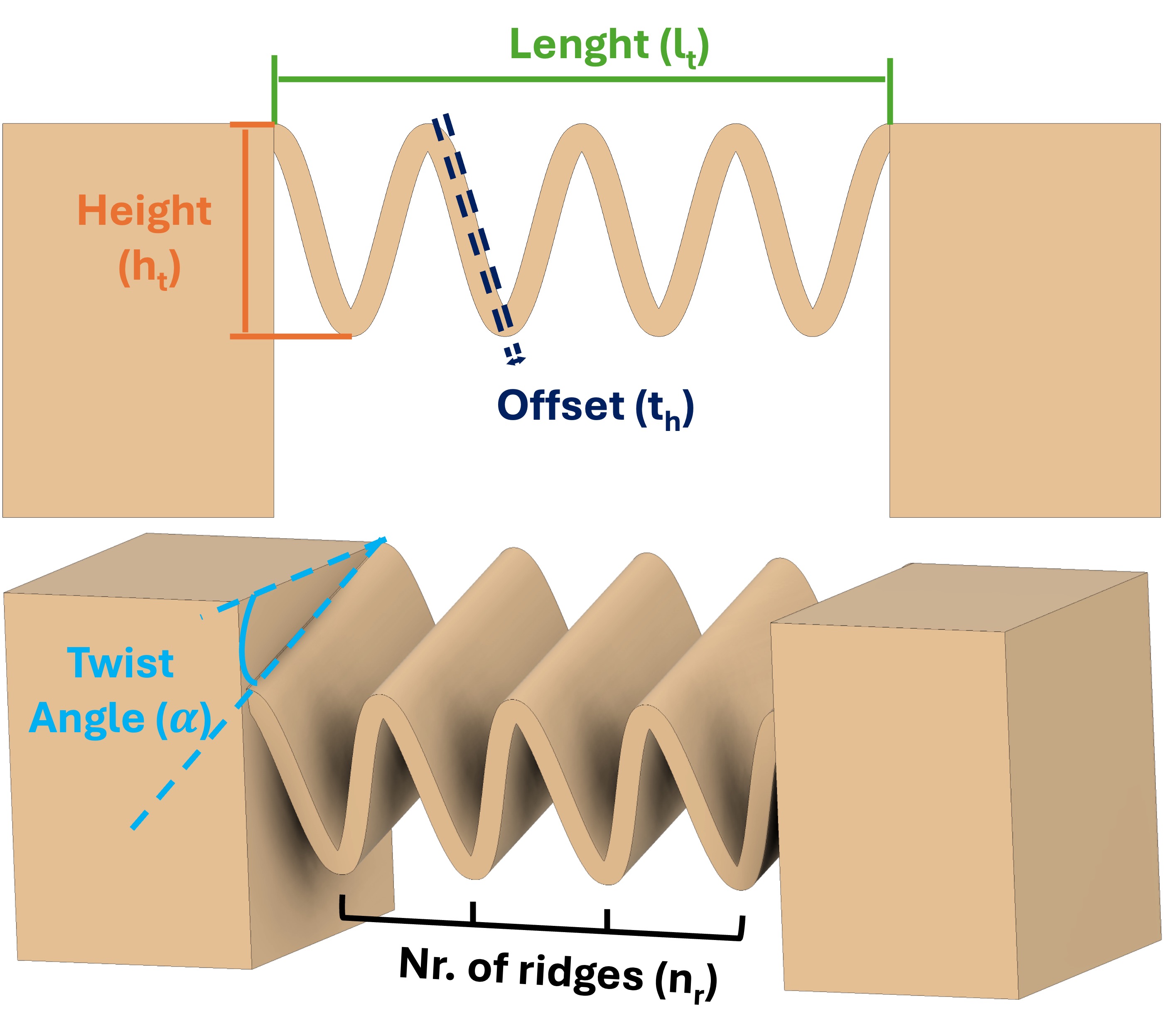}
	\caption{Scheme of the WaveJoint with the parameters that can be varied in the design process based on the desired stiffness.}
	\label{fig:WaveJoint} 
\end{figure}
Dragusanu et al.~\cite{dragusanu2022wavejoints} presented a methodology for designing robotic grippers with 3D-printed, tendon-actuated fingers based on an innovative compliant joint, the WaveJoint (see Figure \ref{fig:WaveJoint}). This approach employs wave-shaped joints to have adjustable finger stiffness that can be modified acting on the wave geometry during design or real-time via pre-compression~\cite{10160353}. 

The correspondence between stiffness values and WaveJoint design parameters is not straightforward. One possible approach to deal with this issue is to estimate a direct mapping training a neural network~\cite{dragusanu2022wavejoints}. However, this approach presents several significant drawbacks: \textit{i}) generating the training data is complex; \textit{ii}) modeling soft materials is challenging, causing simulations to often diverge from actual behavior; \textit{iii}) neural networks generally require large computational resources, increasing design time and complexity.

In this work, we propose an optimization-based approach to directly search for the geometric properties of the WaveJoint that lead to the desired stiffness. This task is modeled as a black-box global optimization problem, and is then handled by means of a surrogate based approach~\cite{gutmann2001radial,costa2018rbfopt} that requires a low number of measurements. Among the advantages of this method we can list the possibility of handling constraints to the geometry of the gripper, of obtaining a solution with a reasonable number of measurements when no data at start and of obtaining, instead, a direct mapping when the number of measurements becomes large enough. Considering a fixed geometric structure (Figure \ref{fig:WaveJoint}), we show that the proposed algorithm is effective at identifying suitable geometric parameters both in simulated and real world tests.

\section{Design of Joint Characteristics by Global Optimization}

\subsection{Problem statement}
A key step in the design of soft-rigid grippers based on WaveJoints amounts to the definition of the flexible joint geometric properties. The correct design of the geometric structure of the joints allows to get suitable stiffness values that lead to the desired fingertip trajectories. For the specific case of WaveJoints, as shown in Figure~\ref{fig:WaveJoint} the geometrical characteristics of the joint are:
\begin{itemize}
	\item $l_t$: joint length - a real number within the interval $[l_\text{min}, l_\text{max}]$;
	\item $n_r$:  number of ridges - an integer between $n_\text{min}$ and $n_\text{max}$;
	\item $h_t$:  wave height (i.e., sinusoidal amplitude) - a real number within the interval $[h_\text{min}, h_\text{max}]$;
	\item $t_h$:  offset (i.e., wave thickness) - a real number within the interval $[t_\text{min}, t_\text{max}]$;
	\item $\alpha$: twist angle - an integer between $\alpha_\text{min}$ and $\alpha_\text{max}$.
\end{itemize}

The problem consists of detecting a set of parameters $(l_t,n_r,h_t,t_h,\alpha)$ that provides a target set of stiffness values $(k_\xi, k_\eta, k_\zeta)$, synthesized on the basis of the gripper specifications.

Let us denote by $\phi:\mathbb{R}^5\to \mathbb{R}^3$ the function mapping the geometric properties of the joints onto the corresponding stiffness values. In order to accomplish the aforementioned task, the approach followed by Dragusanu et al.~\cite{dragusanu2022wavejoints} consisted in approximating the inverse relation $\phi^{-1}$ by means of a feed-forward neural network, trained on a small set of data samples. Once such a network is trained, it can conceptually be used to estimate, for any given triple $(k_\xi, k_\eta, k_\zeta)$, a suitable geometrical configuration of the joint. 
This approach has a number of significant shortcomings:
\begin{enumerate}
	\item \textit{large amounts of data} are required to train neural networks with sufficient expressive power to approximate $\phi^{-1}$ accurately enough;
	\item as \textit{new data} is available, the training process (together with hyperparameters validation) has to be \textit{repeated from scratch}: not only this job is computationally expensive, but it also likely produces a model which is significantly different than the starting one; 
	\item the approach is \textit{incapable of dealing with failure}: assume the configuration of geometric parameters suggested by the neural network model is then found to be wrong, i.e., it actually leads to stiffness values much different than the target ones; then, there is no way of obtaining any kind of second guess;
	\item finally, the parameters $(l_t,n_r,h_t,t_h,\alpha)$ have feasible ranges, and some of them have to take integer values; neural network outputs cannot guarantee these constraints to be satisfied, so that truncation or projection of the retrieved values will often be required in practice; this rounding and readjustment operations are rough and reasonably lead to deteriorating the network performance.
\end{enumerate}

We propose a different approach for the design of joints geometrical properties based on mathematical optimization. The function $\phi$ can be seen as a black box function that can be evaluated, for any setting of $(l_t,n_r,h_t,t_h,\alpha)$, by means of either a structural mechanics analysis based on Finite Element Method (FEM) tools or measurements on a real-world prototype. Given target stiffness values $\hat{k} = (k_\xi, k_\eta, k_\zeta)$, we want to obtain values of $(l_t,n_r,h_t,t_h,\alpha)$ such that:
\begin{itemize}
	\item each parameter takes a value which is physically consistent and implementable;
	\item $\phi(l_t,n_r,h_t,t_h,\alpha) = \hat{k}$.
\end{itemize}
Such values can be obtained as optimal solutions of the following optimization problem
\begin{equation}
	\begin{aligned}
		\label{eq:opt_prob}
		\min_{\substack{l_t,h_c,t_h\in\mathbb{R}\\n_r,\alpha\in\mathbb{Z}}}\;&f(l_t,n_r,h_t,t_h,\alpha)=\|\phi(l_t,n_r,h_t,t_h,\alpha)-\hat{k}\|^2_2\\
		\text{s.t. }\quad& l_\text{min}\le l_t \le l_\text{max}, \quad h_\text{min}\le h_c \le h_\text{max}, \\& 
		t_\text{min}\le t_h \le t_\text{max}, \quad n_\text{min} \le n_r\le n_\text{max}, \\& \alpha_\text{min}\le \alpha \le \alpha_\text{max}.
	\end{aligned}
\end{equation}

Using the terminology from the mathematical optimization literature, problem \eqref{eq:opt_prob} is an instance of global optimization of computationally expensive black--box functions~\cite{cassioli2013global,jones1998efficient}. The key features of this setting are that an analytical expression (and thus derivatives) is not available for the objective function and that its evaluation at some solution has a computational cost that dominates any other operation involved in the optimization process.

Of course, the issue can be generalized to any design process where target properties have to be obtained by properly setting the fundamental characteristics of a mechanical device. 

\subsection{RBF based optimization of expensive black-box functions}
\label{sec:RBF}
To tackle expensive black-box optimization tasks, approaches based on constructing surrogates of the objective function, with increasing accuracy as the number of observations grows, have proved to be solid choices~\cite{jones2001taxonomy}. Briefly, at each iteration, methods of this kind:
\begin{enumerate}[label=(\roman*)]
	\item evaluate the objective function on a new point;
	\item if the solution is satisfactory, the procedure stops;
	\item otherwise, the new observation is added to previous information and used to construct a more accurate approximation (surrogate) of the objective function;
	\item the next point to be evaluated is the global minimizer of an \textit{acquisition function}, i.e., a modified surrogate that privileges unexplored regions of the parameters space. 
\end{enumerate}

A popular approach within this paradigm is that of Radial Basis Function methods~\cite{gutmann2001radial,costa2018rbfopt}. For a formal introduction to this class of methods we refer readers to Appendix \ref{app::A}.

\begin{figure}[tbp]
	\centering
	\includegraphics[width=0.7\textwidth]{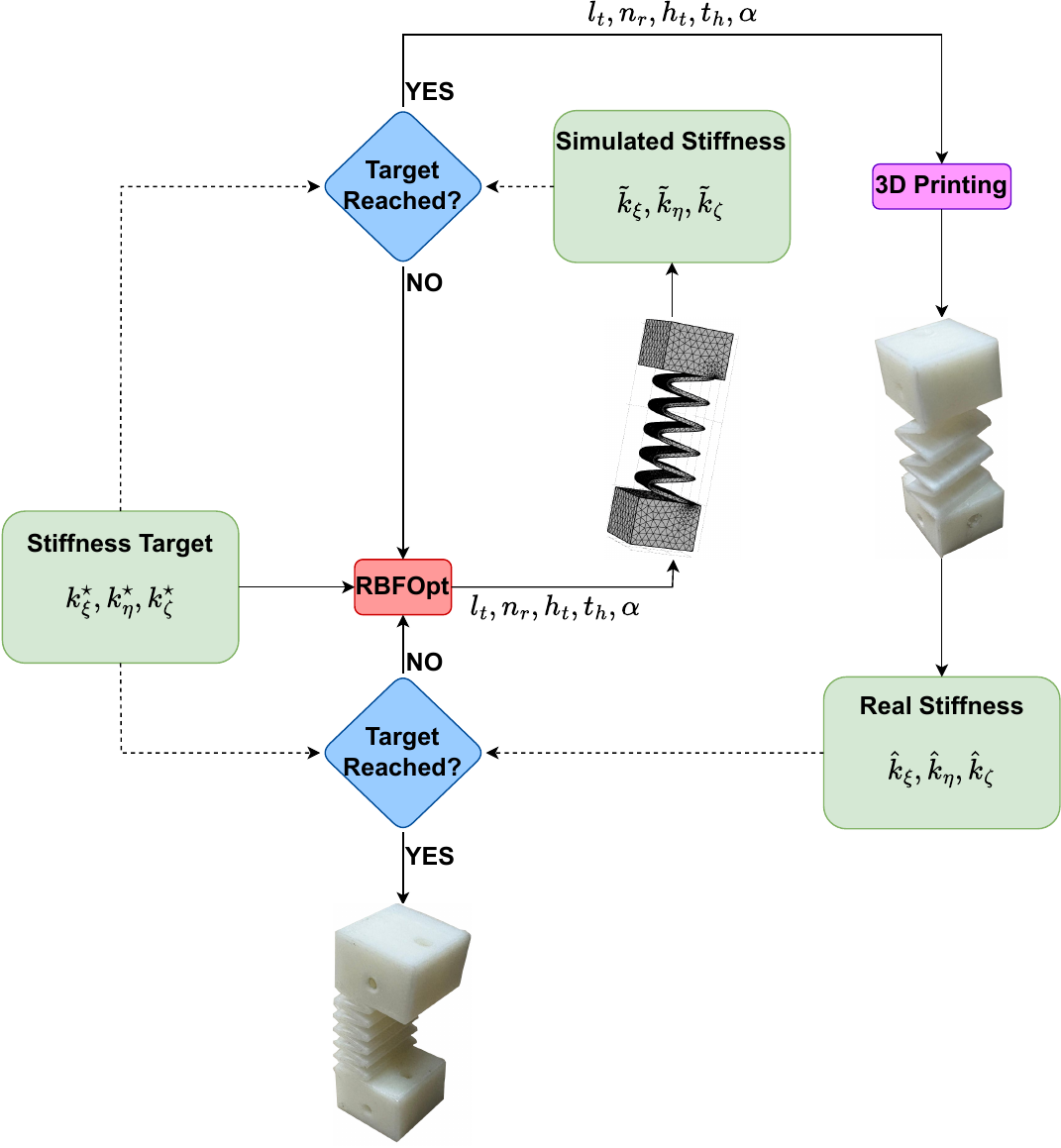}
	\caption{Flowchart of the optimization-based procedure with RBFOpt as black-box optimization software library.}
	\label{fig:flow}
\end{figure}

An  efficient RBF surrogate based algorithm, dealing with expensive black-box optimization problems with bound constraints and possibly integer variables, is available with the recent RBFOpt~\cite{costa2018rbfopt} software library. This software is thus well suited to be used to tackle problem \eqref{eq:opt_prob}. In Figure \ref{fig:flow}, we report a graphical overview of the proposed optimization-based procedure built upon RBFOpt.

This approaches usually require a few tens of function evaluations to reach good quality solutions. Yet, this cost might represent a high computational burden to pay for solving a single gripper design problem, especially if the simulation process were not automated or if we necessarily had to resort to real-world measurements. However, we shall note that, whereas the objective function changes each time we consider a new target $\hat{k}$, the function $\phi$ that we evaluate by expensive simulation or measurement is always the same. Thus, when we want to find the parameters corresponding to $\hat{k}$ by solving problem \eqref{eq:opt_prob}, we can also employ data gathered for past estimation tasks to construct the surrogate, reducing the number of measurements required to determine each single parameter configuration.

This kind of approach overcomes most of the limitations highlighted for the neural networks based methodology; in particular:
\begin{enumerate}
	\item the algorithm becomes \textit{increasingly efficient} with time, since the amount of ``free'' data at the initialization step becomes larger each time an estimation task is solved; 
	\item \textit{no failure} occurs; if the proposed parameters setting is found to be unsuitable to obtain the desired stiffness values, it can be added as a new observation to the interpolation data; this process can be repeated until a satisfactory solution is retrieved;
	\item since we work in the parameters space, we can \textit{explicitly control} the physical \textit{requirements}, treating them as constraints in the optimization problem; all solutions will intrinsically satisfy both bound and integrity constraints.
\end{enumerate}

The methodology proposed here is particularly well suited if the entire process can be automated; this is possible, for example, if the model definition in a simulation environment can be parameterized with respect to the values of the geometric parameters to be estimated - $(l_t,n_r,h_t,t_h,\alpha)$ in the case of WaveJoints. Under this condition, users will be able to carry out a larger number of simulations, i.e., more iterations, and consequently obtain higher quality solutions.

Advanced versions of RBF algorithms - such as the one in RBFOpt library - are also capable of dealing with noise; in particular, RBFOpt is able to distinguish exact observations, assumed to be obtained by a more expensive oracle, and noisy observations from a cheaper oracle. 
This is particularly useful in a setting where we have access to both a simulator and real-world prototypes: the results of a simulation might be imprecise (i.e., noisy), whereas the direct feedback from a real-world prototype should be considered fully reliable; on the other hand, it is obvious that building a prototype is much more expensive than running a simulation. 
The RBFOpt~\cite{costa2018rbfopt} algorithm then operates in such a way that it collects larger amounts of data from the cheaper oracle: these data points, however, are not necessarily interpolated by the surrogate, which rather passes through within confidence interval of the observed noisy value.

\subsection{Incremental Neural Network approach}
If we are open to storing observation results - whether they are from simulations or real-world measurements - and give up having a fixed and definitive network, we might consider adopting an incremental-type scheme, similar to that of RBF approaches, based on neural networks. In particular, one may consider a procedure that, given a set of past observations $\mathcal{D}$ and target values $T$ to obtain with a new configuration $x$, follows these steps: 
\begin{enumerate}[label=(\roman*)]
	\item Train, on dataset $\mathcal{D}$, a neural net $N$ to approximate $\phi^{-1}$; 
	\item Check output $x=N(T)$ of the net given target values $T$;
	\item Measure the true value of $\phi(x)$;
	\item If the value of $\phi(x)$ is close enough to $T$, then stop; otherwise, add the pair $(T,x)$ to the dataset $\mathcal{D}$ and go back to step (i).
\end{enumerate}

As we will show numerically, this approach is actually employable, and shows performance close to those of the RBF optimization approach.
However, contrarily to the latter method, this approach cannot
\begin{itemize}
	\item manage constraints on the values of the parameters $x$ to be estimated;
	\item manage failures possibly occurring due to physically impossible values of $x$ returned by the network as an output; in those cases, it would not be clear where to place the next observation; 
	\item distinguish between exact and noisy observations.
\end{itemize}
Taking into account the above points, the optimization-based approach appears thus to be the most robust methodology, both conceptually and practically. 

\section{Experiments}
Throughout this section, we will require that the geometric parameters of the joint satisfy the following constraints related to physical viability: $l_t \in [15, 30]$, $n_r \in [3, 6]$, $h_c \in [8, 14]$, $t_h \in [0.5, 0.6]$, $\alpha \in [0, 24]$. We carried out all computations on a computer with the following specifics: Ubuntu 22.04, Intel(R) Core(TM) i5-10600KF CPU @ 4.10GHz, 32 GB RAM. The code implementing the optimization process was written in Python3, exploiting the RBFOpt library. For Neural Networks, we employed the \texttt{scikit-learn} Python library. When not carried out on a real-world prototype, the computation of $k = \phi(l_t,n_r,h_c,t_h,\alpha)$ is done by a structural mechanics analysis based on FEM, with simulations done on the COMSOL Multiphysics \textit{6.0} platform, accessed through the MATLAB interface. The parametric design of WaveJoints within COMSOL was done exploiting the following parametric
equation, which describes a cosine function characterized by amplitude, frequency, and phase
\begin{equation}
	\label{eq:wave}
	y(t) = - \left[\left(\frac{\textcolor{orange}{h_t}}{2}-\textcolor{blue}{t_h}\right)cos\left(2\pi\frac{\textcolor{purple}{n_r}}{\textcolor{green}{l_t}}t\right)\right].  
\end{equation}

For the rigid parts of the WaveJoint, a simple geometry with fixed parameters was used, specifically a rectangular parallelepiped with dimensions of $11\times15\times16$ mm. The component was modeled in COMSOL using the Geometry workspace, where two distinct work planes were defined: one for creating the wave using   \eqref{eq:wave} and the other for the two parallelepipeds.
The wave's parameters and the dimensions of the parallelepipeds were defined through the Parameters section in the Global Definitions workspace.
Finally, the 3D component of the WaveJoint was obtained by applying a series of operations: the extrusion of the wave and the parallelepipeds, the rotation of the wave based on $\alpha$, and the union of the three elements into a single body. 

By varying the parameters shown in Figure \ref{fig:WaveJoint}, different configurations of the joint can be achieved. The material was set to ABS (Acrylonitrile Butadiene Styrene) with a Young's modulus \( E = 2.30 \ \text{GPa} \), density \( \rho = 1.06 \ \text{g/cm}^3 \), and Poisson's ratio \( \nu = 0.35 \).
Experiments were carried out setting the mesh element size in COMSOL to \texttt{fine}.

\begin{figure}[tbp]
	\centering
	\subfloat[\label{fig:frontalFEM}]{
		\includegraphics[width=0.3\columnwidth]{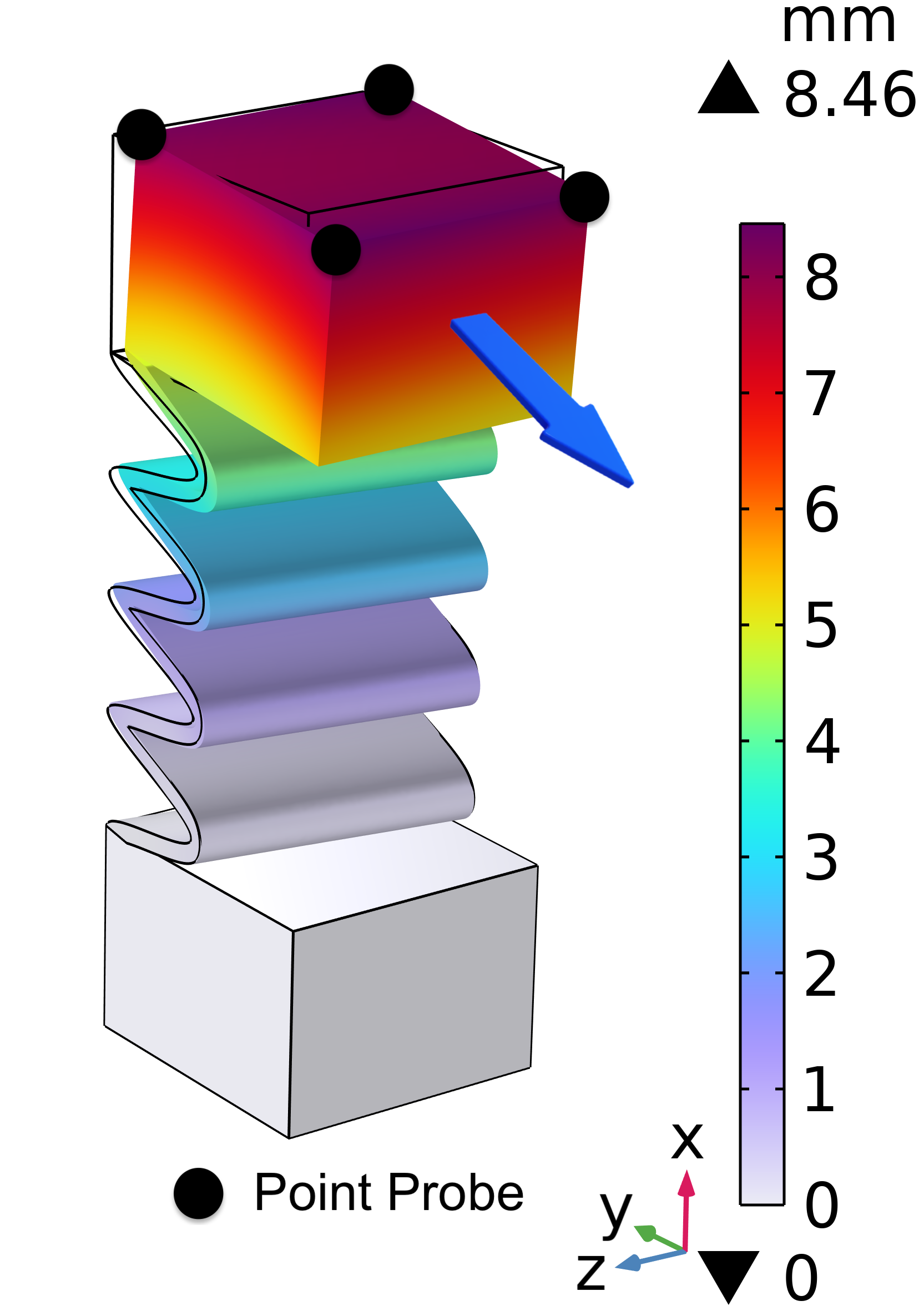} 
	} 
	\subfloat[\label{fig:lateralFEM}]{
		\includegraphics[width=0.321\columnwidth]{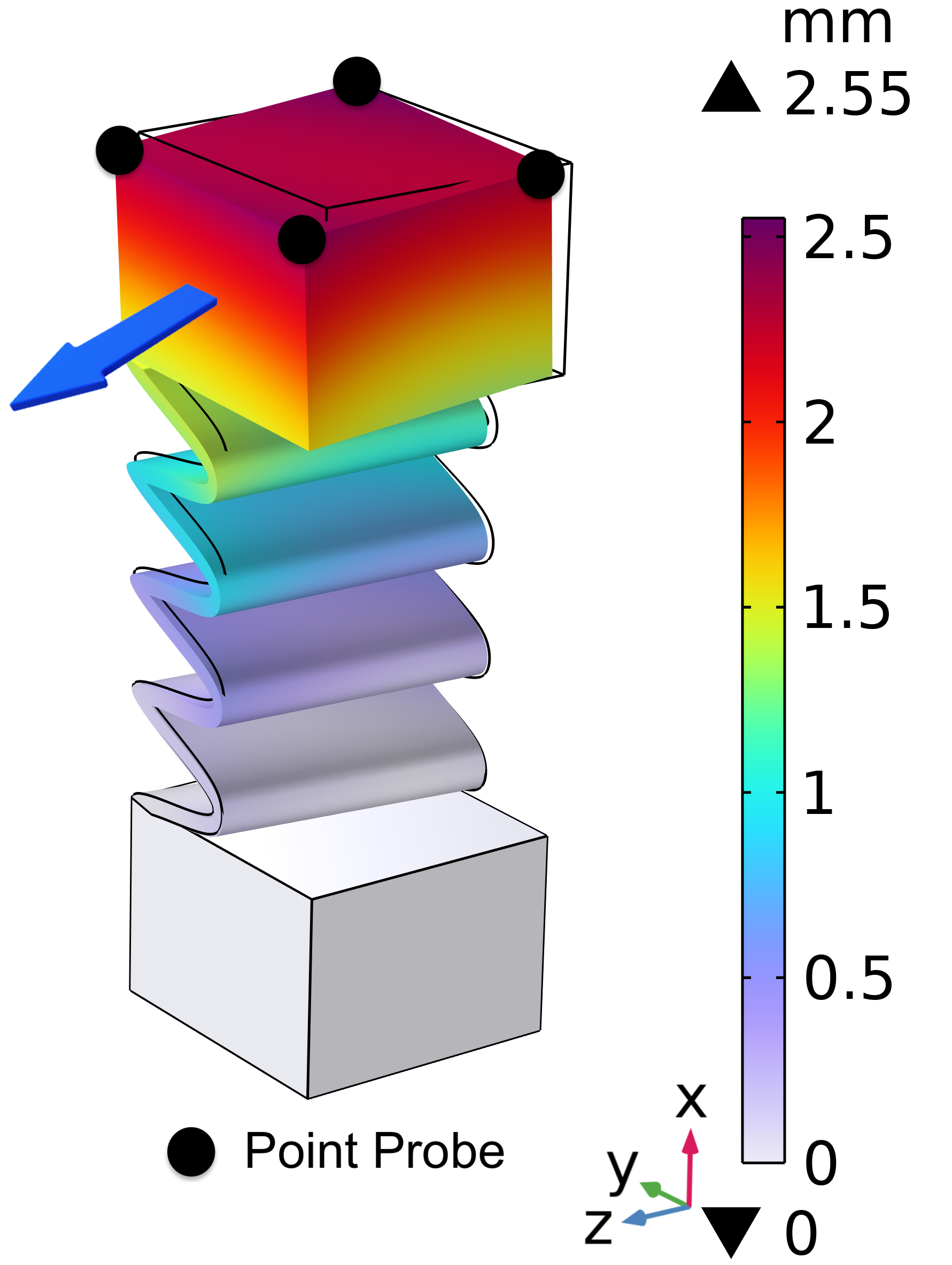}
	}  
	\subfloat[\label{fig:topFEM}]{
		\includegraphics[width=0.31\columnwidth]{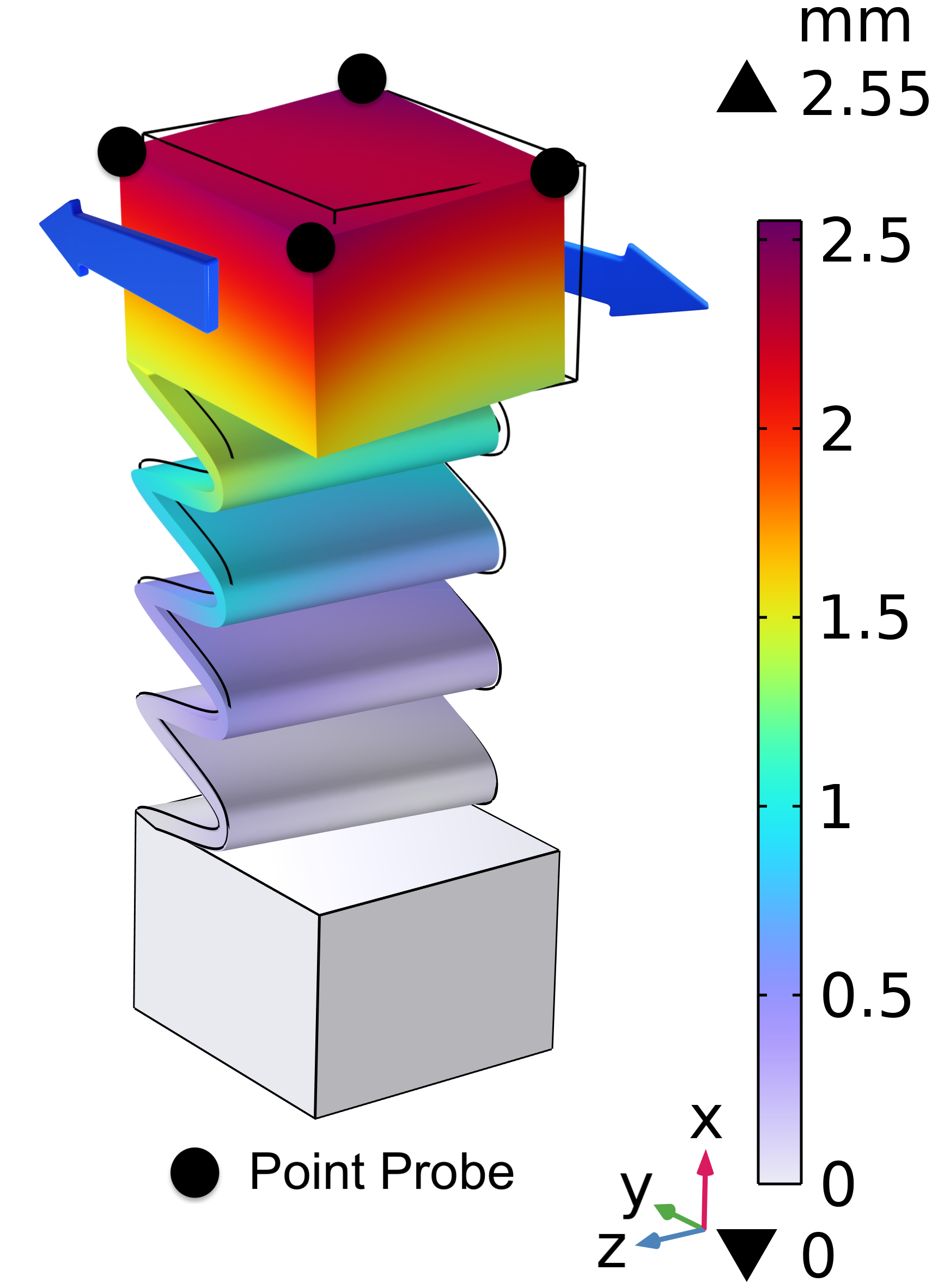}
	}  
	\caption{Results in terms of displacements of a representative simulation for stiffness evaluation based on FEM structural analysis. For each simulation a total of 1 N force was applied with the following parameters of the WaveJoint: \(\textcolor{black}{n_r}=4\), \(\textcolor{green}{l_t}= 25 \text{mm}\), \(\textcolor{blue}{t_h}=0.5\text{mm}\), \(\textcolor{cyan}{\alpha}=10^\circ\), and \(\textcolor{orange}{h_t}=5 \text{mm}\). (a) Longitudinal $(k_\xi)$ bending simulation. (b) Lateral $(k_\eta)$ bending simulation. (c) Torsional $(k_\zeta)$ simulation.}
	\label{fig:FEMComsol}
\end{figure}

Each FEM simulation was evaluated in terms of displacement, using point probes positioned on the parallelepiped vertices (Figure \ref{fig:FEMComsol}). Three loading cases were simulated: bending in the longitudinal plane (Figure \ref{fig:frontalFEM}), bending in the lateral plane (Figure \ref{fig:lateralFEM}), and torsion (Figure \ref{fig:topFEM}). In each case, a force of 1 N was applied. For the torsion force simulation, two equal and opposite forces were applied on the lateral parallelepiped faces, as shown in Figure \ref{fig:topFEM}, making an output momentum of 1 Nmm.
The corresponding stiffness was evaluated based on the final displacement obtained and the parallelepiped geometry. The result, in terms of displacement,  of a representative simulation with \(\textcolor{black}{n_r}=4\), \(\textcolor{green}{l_t}= 25 \text{mm}\), \(\textcolor{blue}{t_h}=0.5\text{mm}\), \(\textcolor{cyan}{\alpha}=10^\circ\), and \(\textcolor{orange}{h_t}=5 \text{mm}\), is shown in Figure \ref{fig:FEMComsol}. Notice that some parameters configurations might lead to COMSOL failures, due to physical values inconsistencies. In these cases, the objective value within RBFOpt is considered to be $+\infty$.

%

\begin{figure}[tbp]
	\centering
	\subfloat[\label{fig:frontalbending}]{
		\includegraphics[width=0.25\columnwidth]{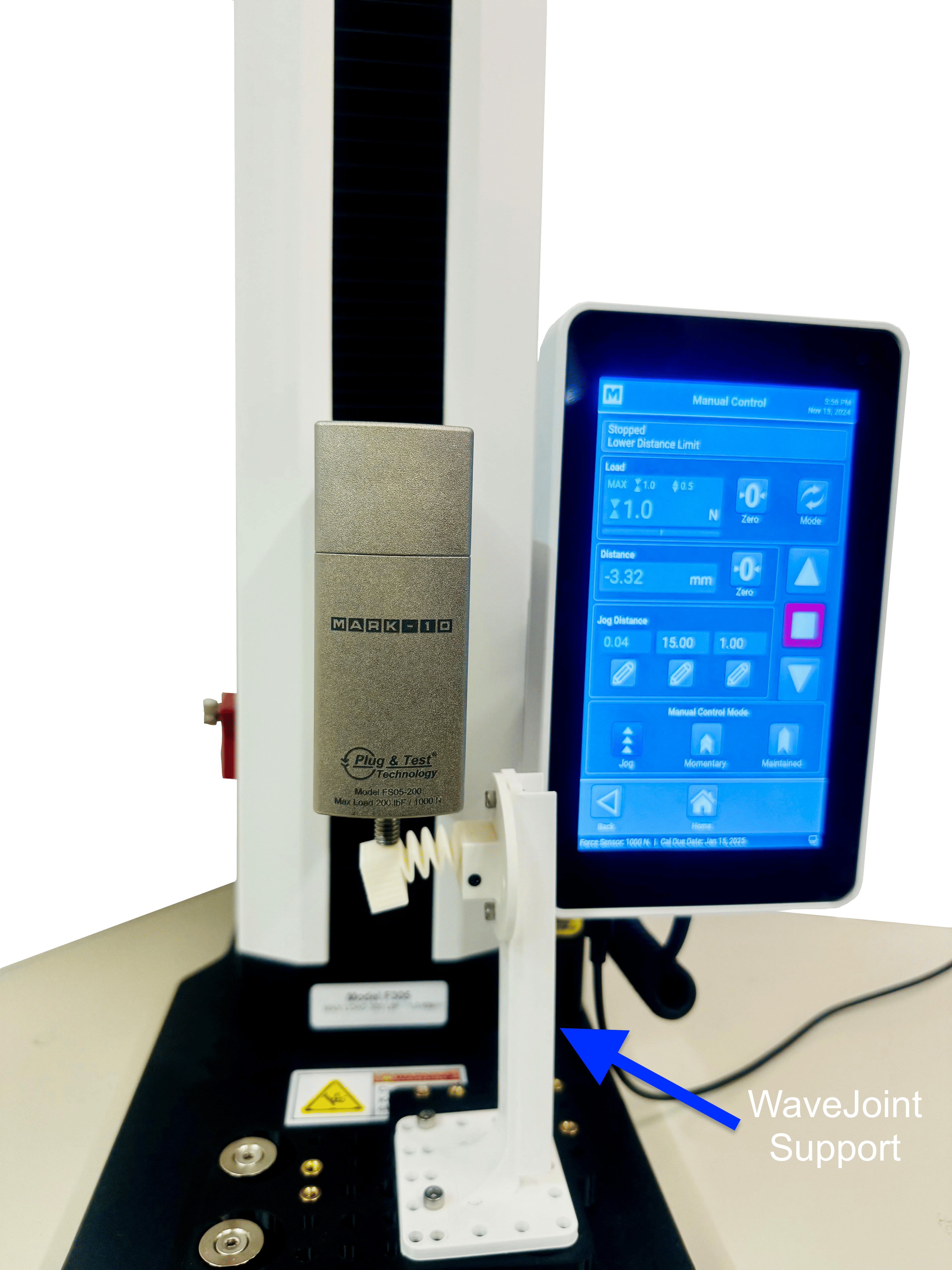} 
	} 
	\hfil
	\subfloat[\label{fig:lateralbending}]{
		\includegraphics[width=0.25\columnwidth]{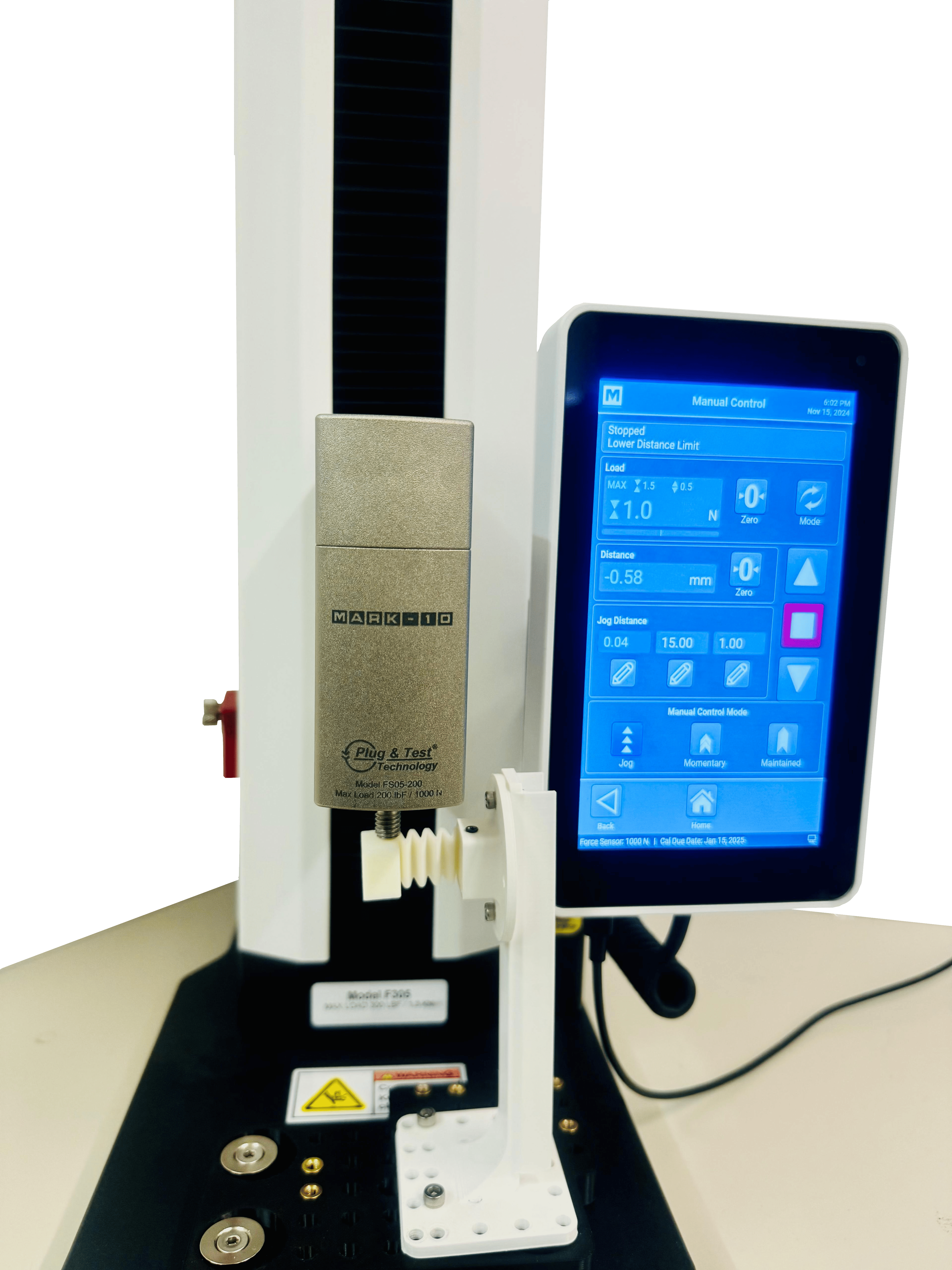}
	}
	\\
	\subfloat[\label{fig:twistall}]{
		\includegraphics[width=0.6\columnwidth]{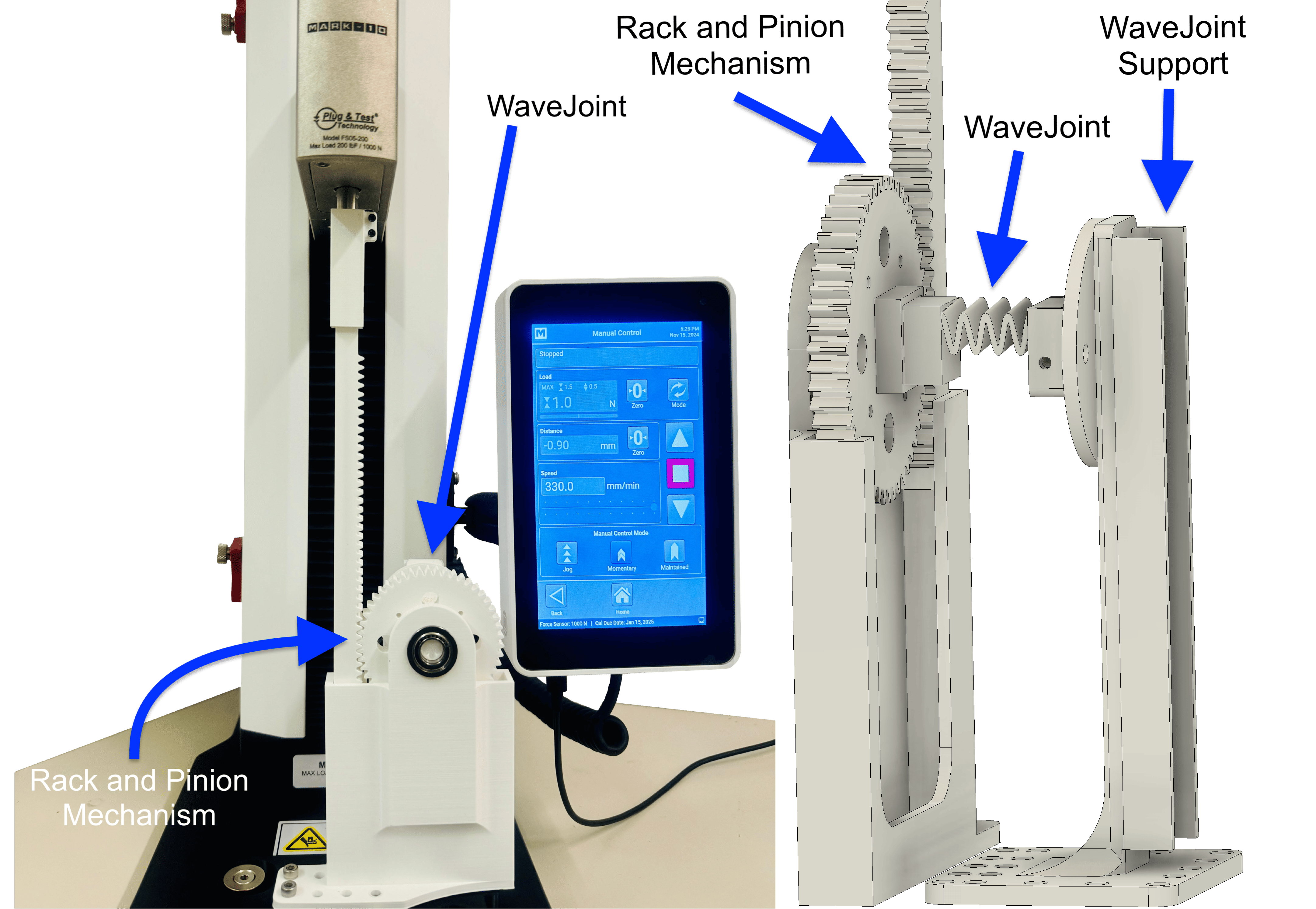}
	}  
	\caption{Test on bench for a WaveJoint sample. (a) Longitudinal $(k_\xi)$ bending test. (b) Lateral $(k_\eta)$ bending test. (c) Torsional $(k_\zeta)$ test.}
	\label{fig:TESTMachine}
\end{figure}

Concerning the physically fabricated joints, all components were realized using FDM (Fused Deposition Modeling) 3D printing with ABS material, the same assumed in simulations. Each sample was experimentally tested using the motorized force test rig  MARK-10 F305 shown in Figure~\ref{fig:TESTMachine}. 
This setup consists of a vertical testing station equipped with a FS05-200 dynamometer and a EMP001-1 distance measurement package. During testing, deformation was induced by a flexion test, where applied forces and corresponding displacements were recorded. The testing force was set to 1 N, consistent with simulation parameters.

To measure deformation along the three axes and calculate joint stiffness, $ k_\xi, k_\eta, k_\zeta $, supports were made. Figures \ref{fig:frontalbending} and \ref{fig:lateralbending} illustrate how the module was evaluated for bending in both the longitudinal $(k_\xi)$ and lateral planes $(k_\eta)$. For torsional testing $(k_\zeta)$, we adapted the linear movement of the test rig to rotational motion through a rack and pinion mechanism, as shown in Figure \ref{fig:twistall}. The pinion gear and the WaveJoint were rigidly connected via a custom housing part on the gear, to ensure the rotational motion of the gear was transferred to the joint. 

For each tested module, the $ k_\xi, k_\eta$, and $k_\zeta $ evaluation involved three distinct tests, each consisting of five consecutive trials. During these trials, the test rig applied a holding test force, and the resulting displacements was recorded. The joint stiffness, along the three axes, was then calculated based on the average displacements values.  


\subsection{Preliminary Experiments}
In a first round of experiments, we set RBFOpt to stop when either a maximum number of iterations of 500 was reached, or a solution with an objective value $<10^{-5}$ was obtained (note that we know that the value of the global optimum is 0).

As a first experiment, we chose different ``Target'' configurations of geometric parameters of the WaveJoint $(l_t,n_r,h_t,t_h,\alpha)$ and computed the corresponding stiffness values $(k_\xi, k_\eta, k_\zeta)$ by FEM simulations in COMSOL. For each of these test cases, we ran RBFOpt to solve the corresponding instance of problem \eqref{eq:opt_prob}. Note that, instead of the simpler objective function from \eqref{eq:opt_prob}, we used the squared norm of relative errors, i.e., $\mathcal{R} = \left\|R\right\|^2_2$,  with $R_i = \frac{\phi_i(x)-T_i}{T_i}$.
%
In the following, we will refer to this quantity as ``residual''. Considering relative errors is beneficial to get uniformly accurate solutions, since the scales of values for $(k_\xi, k_\eta, k_\zeta)$ might be different.

Demanding residuals on the order of $10^{-5}$ may be excessive in real-world applications. Firstly, achieving precision up to several decimal places in stiffness values is often unnecessary to attain the desired physical behavior from the joint. 
Secondly, simulators introduce non-negligible errors compared to real-world observations, rendering the pursuit of such precision from a simulator nonsensical. The chosen value of desired accuracy allows us to evaluate how our approach's precision evolves over time and the extent to which we can advance it in an ideal setting.

\begin{table}[tbp]
	\centering
	\renewcommand{\arraystretch}{1.5}%
	\scriptsize
	\setlength\tabcolsep{2.5pt}
	\caption{Results of WaveJoints design by RBFOpt, according to FEM simulations. On the left: no data at initialization. On the right: 3043 data points at initialization. Further results can be found in Tables \ref{tab:f_before}-\ref{tab:f_after} of Appendix \ref{app::B}.}
	\label{tab::preliminary}
	\begin{minipage}{\textwidth}
		\begin{minipage}{0.49\textwidth}
			\centering
			\begin{tabular}{|c||ccccc||ccc||c||c|}%
				\hline%
				\multirow{2}{*}{Type}&\multicolumn{5}{c||}{Configuration}&\multicolumn{3}{c||}{Stiffness}&\multirow{2}{*}{$\mathcal{R}$}&\multirow{2}{*}{Time(h)}\\%
				\cline{2%
					-%
					9}%
				&$l_t$&$n_r$&$h_t$&$t_h$&$\alpha$&$k_\xi$&$k_\eta$&$k_\zeta$&&\\%
				\hline%
				\hline%
				Target&15&3&8.0&0.5&0&990.3&35.28&5.09&\multirow{2}{*}{5e{-}5}&\multirow{2}{*}{2.59}\\%
				\cline{1%
					-%
					9}%
				Final&15.03&3&8.03&0.5&0&994.77&35.15&5.11&&\\%
				\hline%
				\hline%
				Target&20&3&14&0.5&0&741.28&6&1.97&\multirow{2}{*}{3e{-}5}&\multirow{2}{*}{2.63}\\%
				\cline{1%
					-%
					9}%
				Final&27.95&4&11.66&0.59&11&744.99&6&1.98&&\\%
				\hline%
				\hline%
				Target&20&3&10&0.5&21&635.47&6.22&2.8&\multirow{2}{*}{0}&\multirow{2}{*}{1.1}\\%
				\cline{1%
					-%
					9}%
				Final&27.42&3&10.64&0.6&20&635.91&6.21&2.79&&\\%
				\hline%
			\end{tabular}%
		\end{minipage}
		\hfill
		\begin{minipage}{0.49\textwidth}
			\centering
			\begin{tabular}{|c||ccccc||ccc||c||c|}%
				\hline%
				\multirow{2}{*}{Type}&\multicolumn{5}{c||}{Configuration}&\multicolumn{3}{c||}{Stiffness}&\multirow{2}{*}{$\mathcal{R}$}&\multirow{2}{*}{Time(h)}\\%
				\cline{2%
					-%
					9}%
				&$l_t$&$n_r$&$h_t$&$t_h$&$\alpha$&$k_\xi$&$k_\eta$&$k_\zeta$&&\\%
				\hline%
				\hline%
				Target&20&4&10&0.5&0&803.39&12.49&2.38&\multirow{2}{*}{8e{-}4}&\multirow{2}{*}{4.89}\\%
				\cline{1%
					-%
					9}%
				Final&27.05&5&9.08&0.58&5&818.76&12.7&2.34&&\\%
				\hline%
				\hline%
				Target&20&5&10&0.5&6&932.27&14.26&2.45&\multirow{2}{*}{3e{-}5}&\multirow{2}{*}{4.79}\\%
				\cline{1%
					-%
					9}%
				Final&27.86&6&8.99&0.6&7&936.3&14.22&2.45&&\\%
				\hline%
				\hline%
				Target&20&3&10&0.5&24&630.04&5.73&2.86&\multirow{2}{*}{1e{-}5}&\multirow{2}{*}{4.61}\\%
				\cline{1%
					-%
					9}%
				Final&20.01&3&9.98&0.5&24&629.24&5.72&2.87&&\\%
				\hline%
			\end{tabular}%
		\end{minipage}
	\end{minipage}
\end{table}

The experiment was repeated for 7 problem instances. We report in Table \ref{tab::preliminary} (left) the results for 3 representative cases, the full table can be found in Appendix \ref{app::B}. We can look at the final residual values to assess the quality of the obtained solutions.  The process always ended with stiffness values very close to the target ones. Interestingly, the returned configuration of geometric parameters is equal to the target one only in 1 case out of 7. It is in fact not surprising that different geometric settings of the joint lead to equal stiffness values ($\phi$ is not injective, as it maps points to a lower dimensional space).

We then tested the case where past simulations results are available when we begin the optimization process for the design of a new joint. In particular, the 3043 simulations carried out in the first phase were used for initializing the process in 7 new instances. The full corresponding results can be found in Appendix \ref{app::B}; here we show in Table \ref{tab::preliminary} (right) 3 representative cases. As expected, solutions get more accurate; the runtimes however seem to double, even though the budget of function evaluations (i.e., simulations) is still 500. Indeed, constructing and then optimizing a surrogate based on thousands nodes is no more computationally negligible and the runtime is no more completely dominated by COMSOL simulations. Arguably, a good compromise might consist of only using for the initialization of the algorithm a subset of all the available past observations. 

Employing available information to start the optimization process have beneficial effects on the cost-quality trade-off: the surrogate function shall provide a nicer proxy for the true objective function from early iterations, so that the error might be brought sufficiently down in far fewer steps. 

\begin{figure}[tbp]
	\centering
	\subfloat[]{\includegraphics[width=0.49\textwidth]{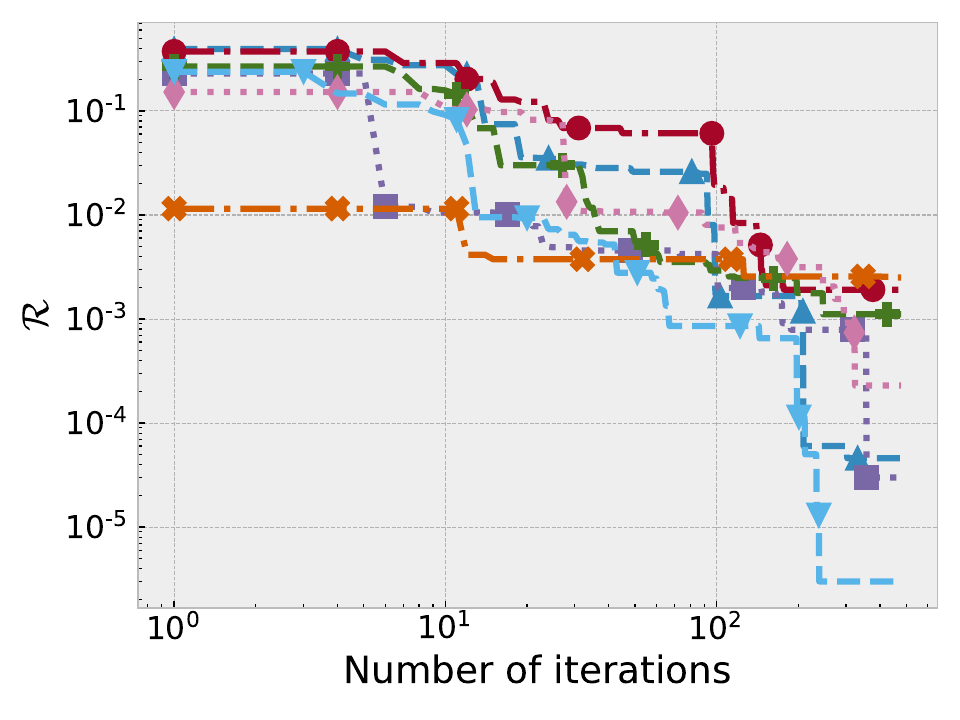}\label{fig:f_a}}
	\hfil
	\subfloat[]{\includegraphics[width=0.49\textwidth]{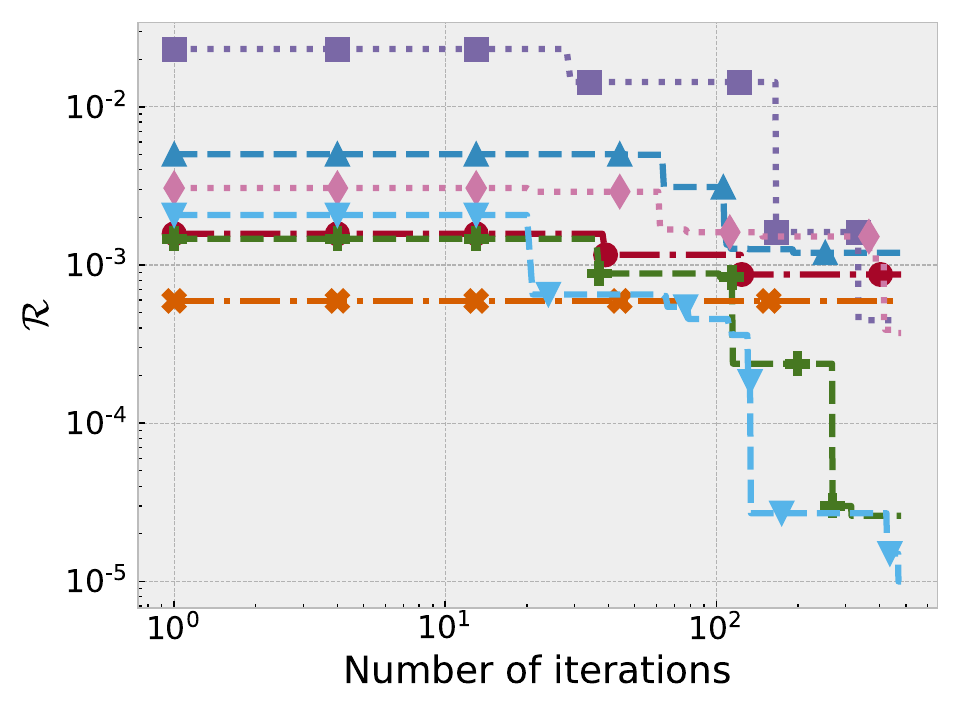}\label{fig:f_b}}
	\caption{Progress of residual along RBFOpt iterations for the 14 problem instances related to Table \ref{tab::preliminary}. (a) 7 instances -- No data point provided at RBFOpt initialization. (b) 7 instances -- 3043 data points provided at RBFOpt initialization.}
	\label{Fig:f}
\end{figure}

To analyze this aspect, we report in Figure \ref{Fig:f} the trend of residuals as new simulations are carried out, for each test instance (note log scale for both axes). When no observation is initially fed to the RBFOpt algorithm (Figure \ref{fig:f_a}), the process naturally starts with larger errors, and acceptable solutions are only reached after a few dozen simulations, while hundreds of iterations are often necessary to reach a higher accuracy. When past data samples are used in the initialization phase, the initial residual is often well below $10^{-2}$, indicating that the first proposed solution might already be acceptable; moreover, the error can be brought down below $10^{-3}$ with a number of iterations ranging from few tens to a hundred.

Accuracy up to $10^{-2}$-$10^{-3}$ might be sufficient in most robotics applications. When sufficiently large amounts of data have been collected, the design problem might be possibly considered solved and no additional attempts might be required before finding a suitable solution. This issue is investigated in the next section.

\subsection{Evaluation of zero-shot approaches}
\label{sec:zero-shot}
In this section we verify if, and eventually at what point, incremental approaches are no more necessary to find suitable configurations for WaveJoints. We verify the behavior, as the number of the available past observations grows, of two ``\textit{zero-shot}'' approaches that do not require users to carry out additional simulations/measurements: 
\begin{itemize}
	\item the neural network approach that was originally employed by Dragusanu et al.~\cite{dragusanu2022wavejoints};
	\item a nearest-neighbor greedy approach that, given a target $k$, returns the configuration in the available dataset with the closest corresponding values of stiffness.
\end{itemize}

\begin{figure*}[tbp]
	\centering
	\subfloat[]{\includegraphics[width=0.32\textwidth]{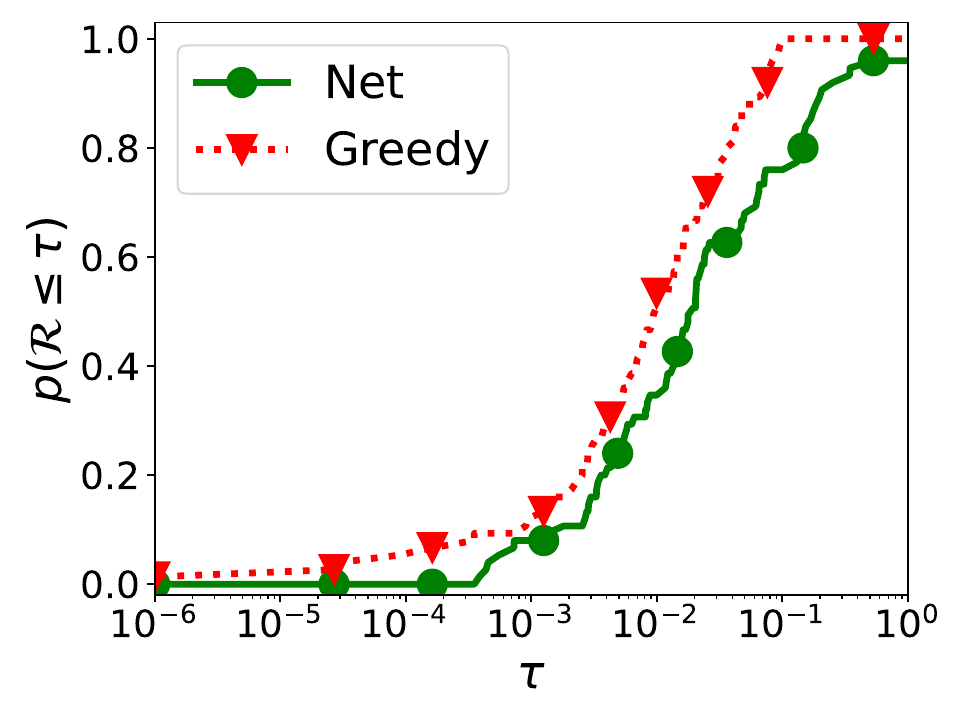}\label{fig:eunet_a}}
	\hfil
	\subfloat[]{\includegraphics[width=0.32\textwidth]{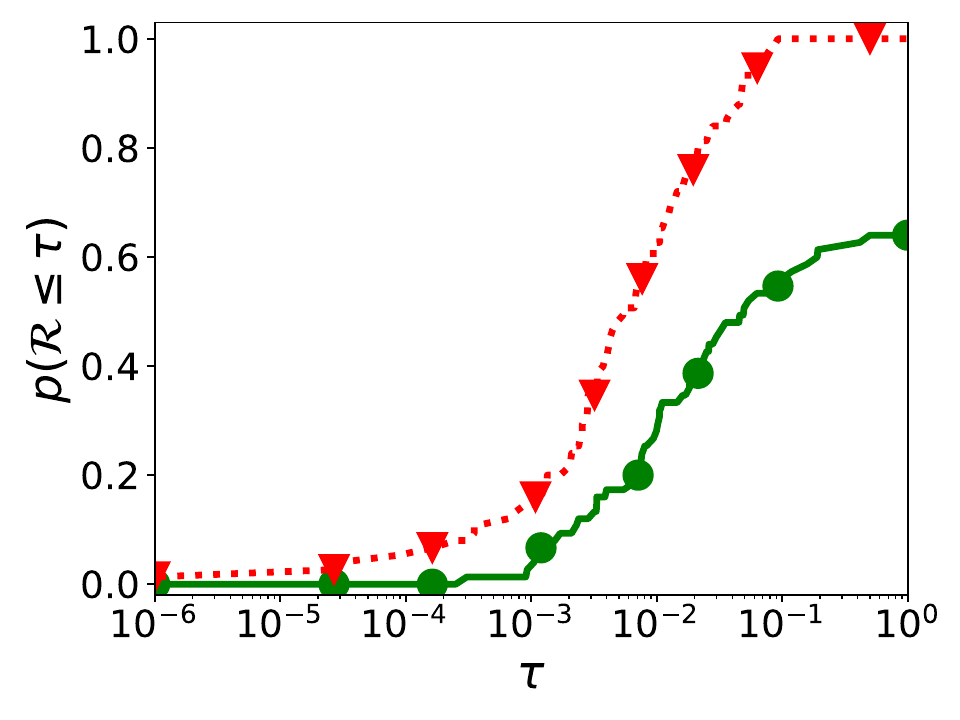}\label{fig:eunet_b}}
	\hfil
	\subfloat[]{\includegraphics[width=0.32\textwidth]{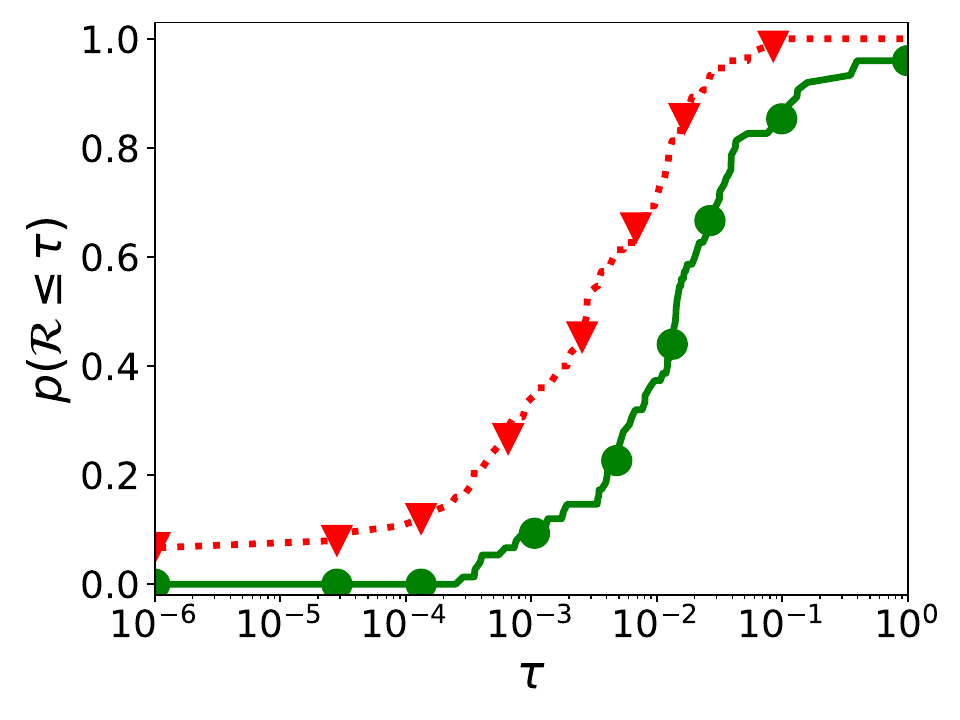}\label{fig:eunet_c}}
	\caption{Cumulative distribution of residuals attained by zero-shot methods over 25 different targets, for different dataset sizes. (a) 1000 observations. (b) 2000 observations. (c) 6000 observations.}
	\label{Fig:eunet}
\end{figure*}

We report in Figure \ref{Fig:eunet} the cumulative distribution of residuals obtained by the two zero-shot approaches upon a benchmark of 25 WaveJoint design problems, given sets of available past observations of 1000, 2000 and 6000 samples respectively.

The nearest-neighbor heuristic exhibits a consistently more robust behavior than the neural network. The latter method, however, still appears to work decently if it is provided with enough training data. 
In absolute terms, examining the results of the nearest-neighbor heuristic, we observe that with 6000 examples in the dataset, very high levels of accuracy are frequently achieved, with very few instances where the residual values are likely unacceptable. With 2000 examples, there are fewer cases of very high accuracy levels; nevertheless, the methodology consistently yields acceptable solutions. However, with only 1000 data points, the approach appears somewhat less reliable.

We might conclude that, when a few thousands samples have been collected, the estimation problem could be considered solved employing one of the aforementioned methods. Our interest, for the rest of the Section, will thus lie in the case where few, or even zero information is available at the beginning.

\subsection{Incremental approaches in fully simulated setup}
\label{sec:exp_c}
We report the results of experiments carried out under two main assumptions:
\begin{itemize}
	\item few amounts of data are available at the start of the design process;
	\item we work in a fully simulated setting: calculations are only done by means of the simulator, the quality of the solution is measured based on the output of the simulator; numerical errors introduced by the simulating environment w.r.t.\ the physical world are thus ignored. 
\end{itemize}

We distinguish four scenarios, based on the number of past observations available at the initialization of the process: 25, 50, 100 and 200 samples. For each scenario, we will consider 15 problem instances.
We consider the RBFOpt algorithm with a budget of 5, 10 and 20 additional function evaluations (i.e., simulations) to carry out. We also do the same for the incremental-Net approach.

\begin{figure}[tbp]
	\centering
	\subfloat[\label{fig::rbf-base}]{\includegraphics[width=0.49\textwidth]{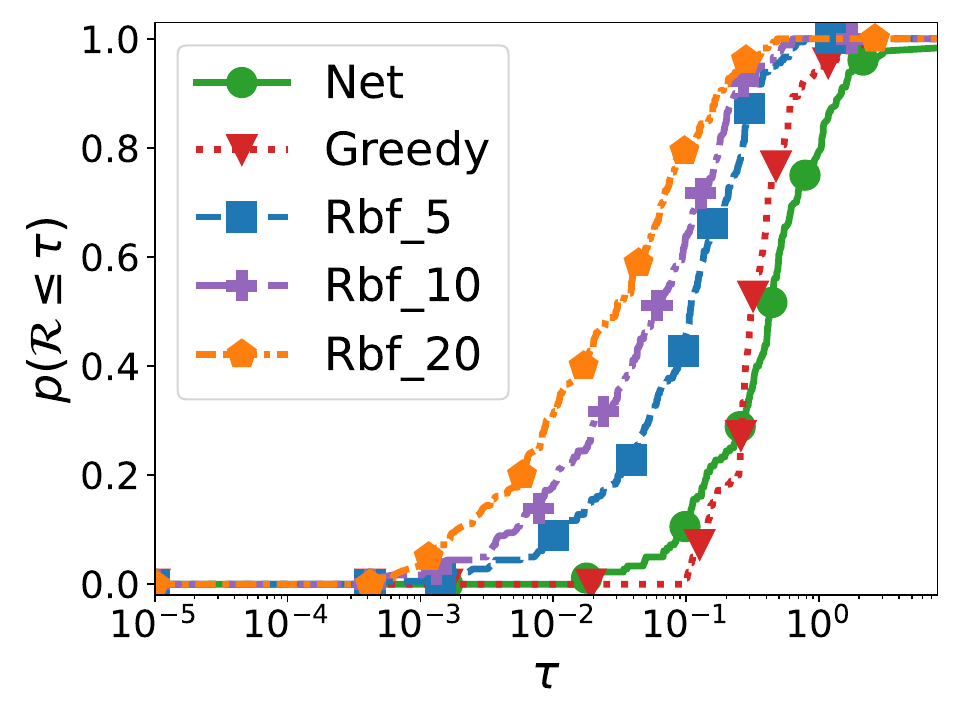}\label{fig:_inc_vs_bench_a}}
	\hfil
	\subfloat[\label{fig::inc-base}]{\includegraphics[width=0.49\textwidth]{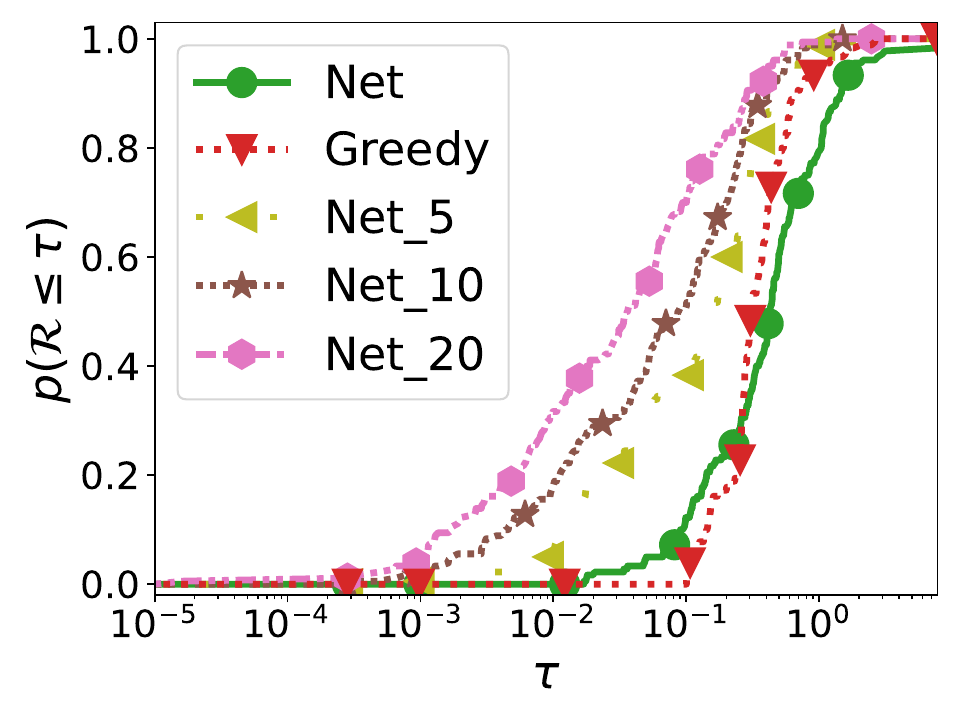}\label{fig:_inc_vs_bench_b}}
	\caption{Cumulative distribution of residuals attained upon a benchmark of 60 instances with 25, 50, 100 and 200 data points at initialization. (a) RBFOpt vs.\ baselines (b) Incremental-Net vs.\ baselines}
	\label{Fig:inc_vs_bench}
\end{figure}

First, we show the results, in the form of cumulative distribution of final residual, upon the entire benchmark of 15 problems for the RBFOpt approaches (Figure \ref{fig::rbf-base}) and the incremental-Net approaches (Figure \ref{fig::inc-base}). In both figures, we also report the performance of the two zero-shot approaches analyzed in Section \ref{sec:zero-shot}. 

As expected, the incremental approaches lead to improved solutions w.r.t.\ the zero-shot baselines. Then, the other expected yet notable result is that providing the incremental approaches with a larger budget of function evaluations allows to improve the robustness of the method. 
Finally, while few tens of data points are not enough to produce high accuracy solutions, we can see that acceptable levels of residuals can be attained consistently with the incremental approaches. This aspect is 
the most important one, recalling that high precision is usually not needed in robotics applications, especially with simulated results. 

\begin{figure*}[tbp]
	\centering
	\subfloat[]{\includegraphics[width=0.24\textwidth]{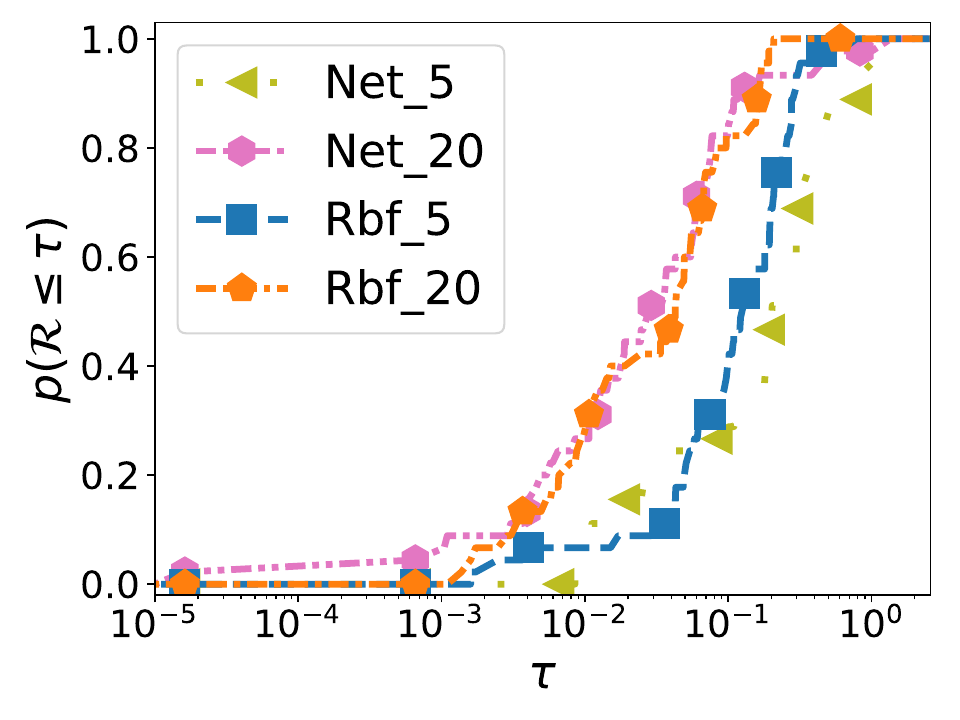}\label{fig:incs_a}}
	\hfil
	\subfloat[]{\includegraphics[width=0.24\textwidth]{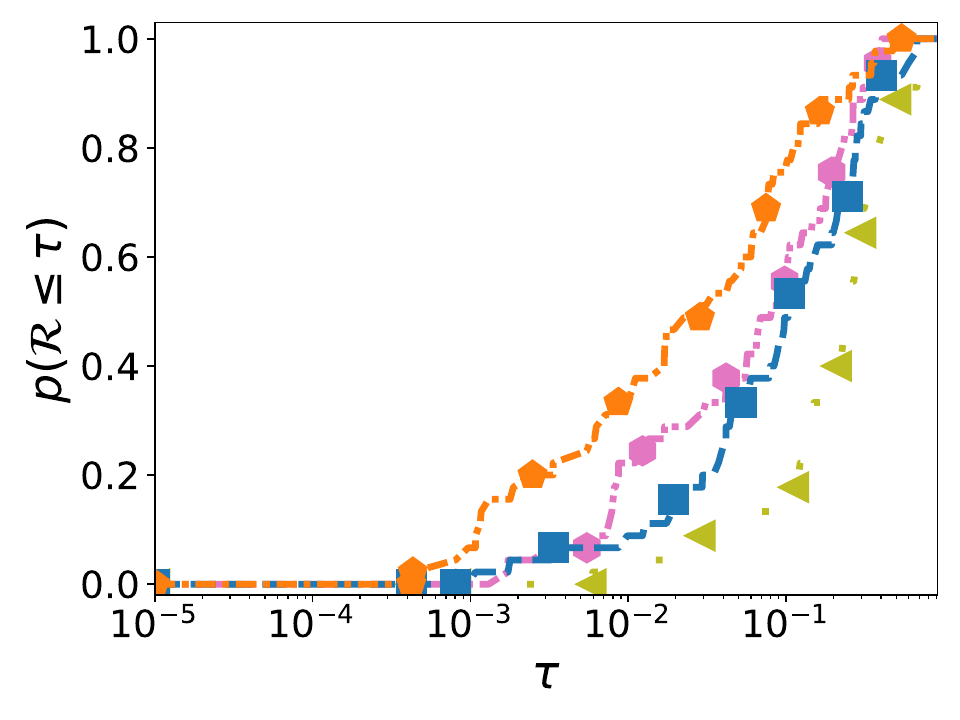}\label{fig:incs_b}}
	\hfil
	\subfloat[]{\includegraphics[width=0.24\textwidth]{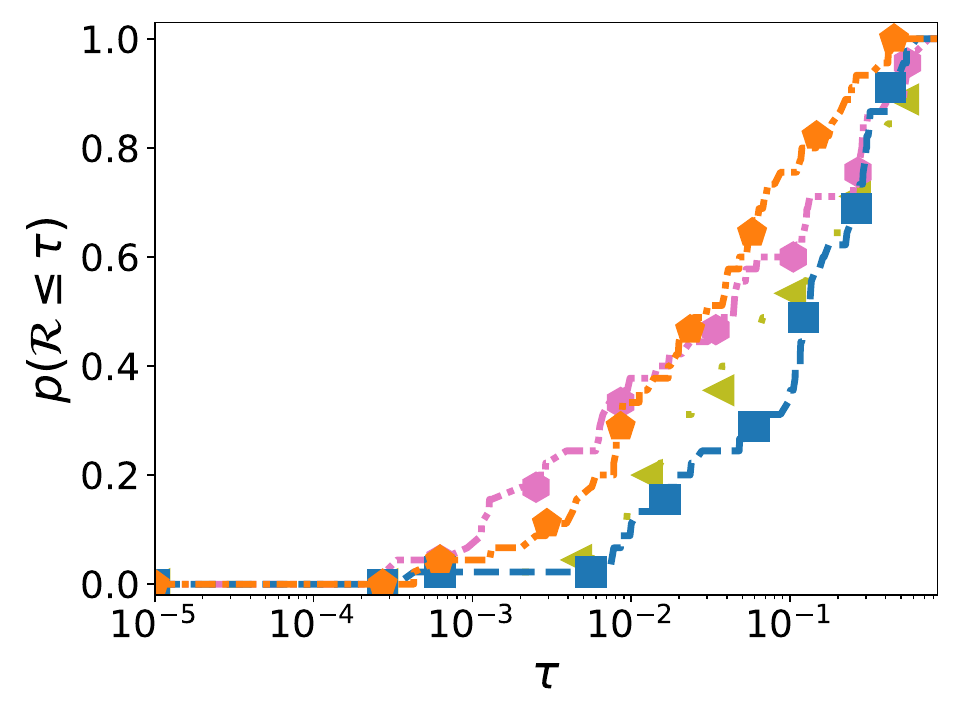}\label{fig:incs_c}}
	\hfil
	\subfloat[]{\includegraphics[width=0.24\textwidth]{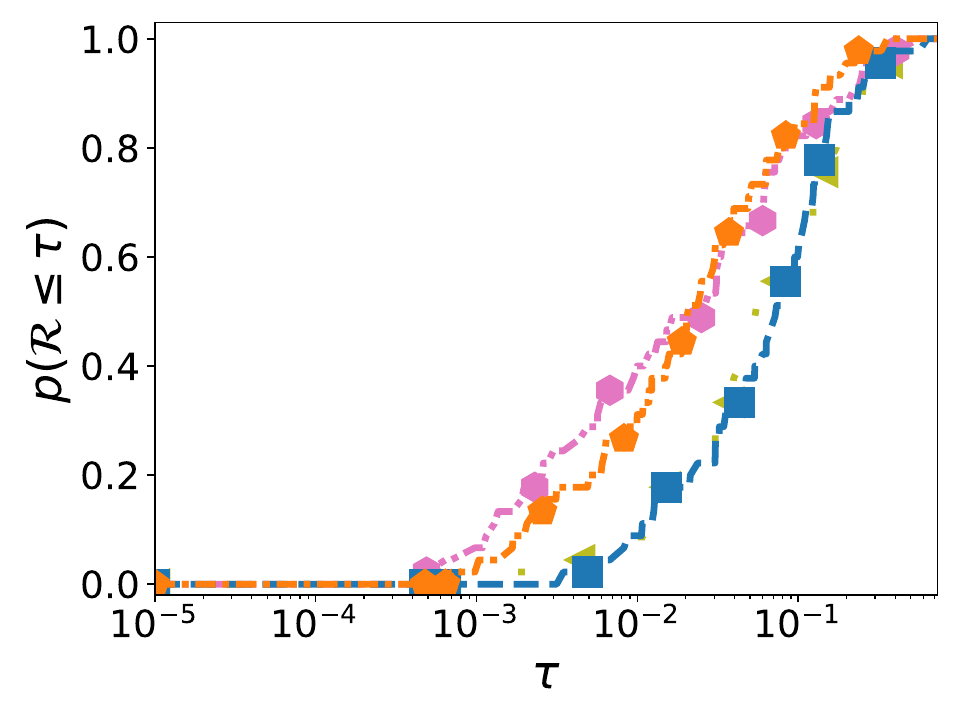}\label{fig:incs_d}}
	\caption{Cumulative distribution of residuals attained upon 4 benchmarks of 15 instances each. (a) 25 data points at initialization (b) 50 data points at initialization (c) 100 data points at initialization (d) 200 data points at initialization}
	\label{fig:incs}
\end{figure*}

In Figure \ref{fig:incs}, more detailed results for the four different scenarios were provided showing a direct comparison between RBFOpt and incremental-Net approaches (we only report the cases with a budget of 5 and 20 functions evaluations for the sake of clarity). 
The performance of the two different methodologies are comparable; there seems to be an interesting, slight advantage of the RBF optimization approach in terms of robustness, as in each of the 4 cases we can observe that the RBF curves are above their incremental-Net counterparts on the right end of the plot.

\subsection{Real world measures and performance of the noisy RBF approach}

The results from Section \ref{sec:exp_c} would make us confident about the proposed approach being a solid choice for the design of devices with target properties, if only simulators perfectly reproduced real-world behavior. Of course, this is not the case; we thus validate the approach on a realistic case study. We have to set the values of $l_t,n_r,h_c,t,\alpha$ for a WaveJoint so that the real-world measures of stiffness values will be close enough to some targets. We assume we have no prior information about the problem; in other words, we address the task starting from scratch, as it usually happens in applications. 

To deal with both simulated and real-world measures obtained by 3D-printing a module and evaluating it on the test rig, we exploited the option (mentioned in \ref{sec:RBF}) of RBFOpt for handle noise: real-world measures can be considered observations from an expensive exact oracle, whereas simulated measures can be set as observations from a cheap, yet noisy oracle. The RBFOpt solver is informed about cheap observations being noisy, with a noise cap that, based on numerous empirical tests, was set to 30\% relative error.

We ran the RBFOpt based procedure to find a configuration leading to stiffness values close to $(k_\xi, k_\eta, k_\zeta) = (1238.27, 50.84, 5)$. We considered it reasonable to carry out 25 simulations between two real-world measures. After 25 new simulated, noisy observations, the algorithm asks to check the real-world measure corresponding to the best value obtained so far by the simulator;  if the obtained module is considered not satisfactory from an engineering standpoint, the exact measure is added to the set of noiseless observations used by RBFOpt to build the surrogate objective function. In Table \ref{tab:incremental}, we show the stiffness values (both according to the simulator and the empirical measure) of the modules that have been 3D-printed throughout the process. We can observe that the first two attempts are far from the target, because RBFOpt is still exploring parts of the space related to physically extreme configurations where there is a large gap between the simulated and the true measure. After just four attempts the proposed solution starts being close to the target and the residual keeps getting reduced until $\mathcal{R} = 0.13$. At that point, we have reached a reasonably accurate solution, and arguably we cannot hope to go any further due to unavoidable imperfections both in 3D-printing and forces measures.

\begin{table}[tbp]
	\centering
	\renewcommand{\arraystretch}{1.2}%
	\caption{Evolution of RBFOpt process at ``milestone'' iterations where a module is physically 3D-printed. The target stiffness values are $(k_\xi, k_\eta, k_\zeta) = (1238.27, 50.84, 5)$. Information for printed modules n.\ 3,5,6,7 can be found in the Table \ref{tab:incremental_app} of Appendix \ref{app::B}.}
	\label{tab:incremental}
	\scriptsize
	\begin{tabular}{|c|c||c|ccc|c|}%
		\hline%
		\multirow{2}{*}{N$^{\circ}$ Simul.}&\multirow{2}{*}{N$^{\circ}$ Real}&\multirow{2}{*}{Type}&\multicolumn{3}{c|}{Stiffness}&\multirow{2}{*}{$\mathcal{R}$}\\%
		\cline{4%
			-%
			6}%
		&&&$k_\xi$&$k_\eta$&$k_\zeta$&\\%
		\hline%
		\hline%
		\multirow{2}{*}{25}&\multirow{2}{*}{1}&Simul.&1652.89&31.07&4.63&0.268979\\%
		\cline{3%
			-%
			7}%
		&&Real&2683.5&47.46&11.25&2.922401\\%
		\hline%
		\hline%
		\multirow{2}{*}{50}&\multirow{2}{*}{2}&Simul.&1590.97&25.5&4.3&0.349188\\%
		\cline{3%
			-%
			7}%
		&&Real&1451.7&23.89&11.29&1.887757\\%
		\hline%
		\hline%
		\multirow{2}{*}{100}&\multirow{2}{*}{4}&Simul.&1641.87&30.18&5.16&0.272236\\%
		\cline{3%
			-%
			7}%
		&&Real&1383.7&30.43&6.21&0.232495\\%
		\hline%
		\hline%
		\multirow{2}{*}{200}&\multirow{2}{*}{8}&Simul.&1353.96&37.0&5.56&0.09524\\%
		\cline{3%
			-%
			7}%
		&&Real&1118.4&32.85&5.05&0.134661\\%
		\hline%
	\end{tabular}%
\end{table}

We finally report the result of an additional experiment, carried out to investigate the extreme case where we are willing to only build one module, at the end of the process, after a reasonable number of simulations is carried out. In other words, we test how the approach purely based on simulations performs in practice. We considered a scenario with target $(k_\xi, k_\eta, k_\zeta) = (982.36, 37.87, 5.87)$ and with a budget of 170 simulations. The final result gave simulated stiffness values of $(1037.32, 35.9, 5.53)$ - $\mathcal{R} = 0.009$ - translating to real-world stiffness measures of $(1054.2, 33.33, 5.94)$, with $\mathcal{R} = 0.01$, that are acceptable for design purposes.

\subsection{Study of obtainable stiffness values}
Our approach interestingly allows also to evaluate the space of stiffness values and the region of configurations that are obtainable with physically realizable WaveJoints. Using the approach proposed in this work with a target configuration that is actually unreachable, we observe that after a large number of RBFOpt iterations the error remains stable on high values, even if the parameters space has been thoroughly explored.

\begin{figure}[tbp]
	\centering
	\subfloat[]{\includegraphics[width=0.3\textwidth]{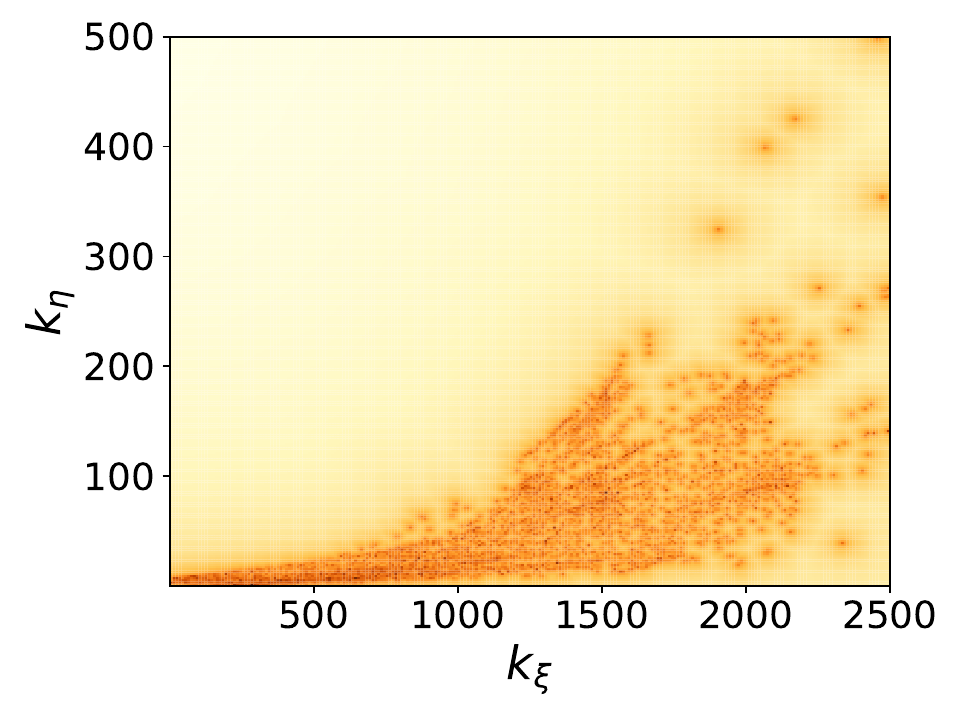}}
	\hfil
	\subfloat[]{\includegraphics[width=0.3\textwidth]{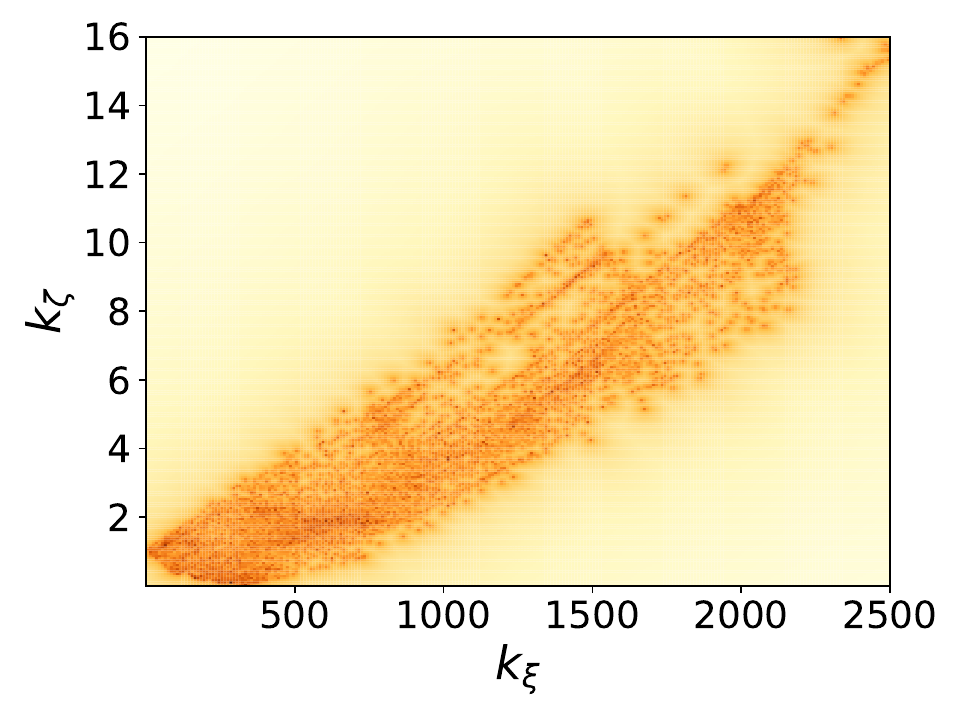}}
	\hfil
	\subfloat[]{\includegraphics[width=0.3\textwidth]{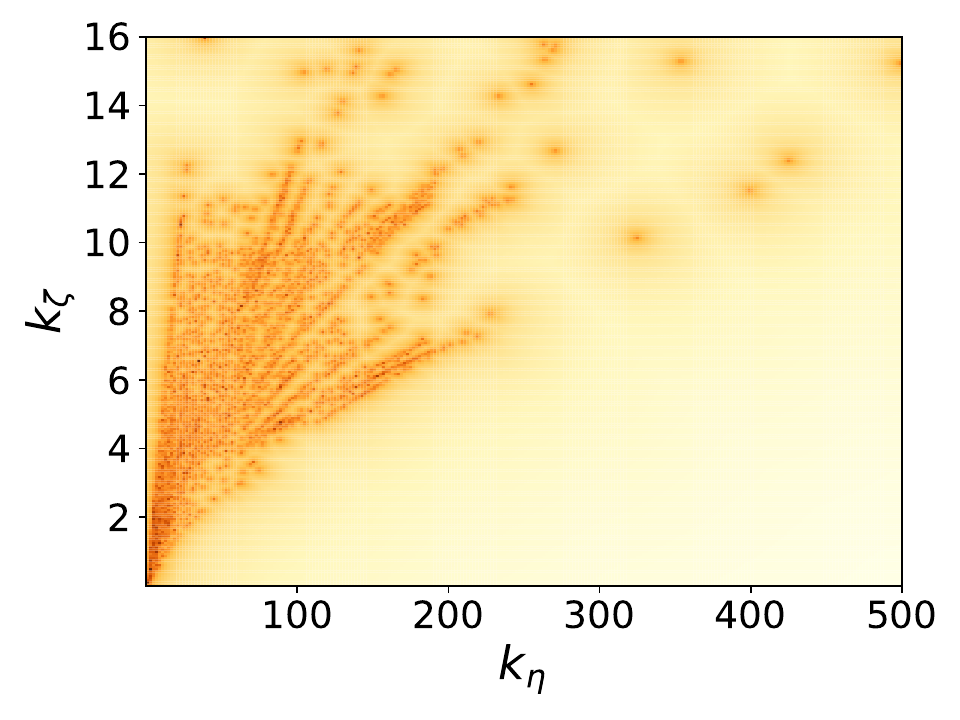}}
	\hfil
	\subfloat{\includegraphics[width=0.0225\textwidth]{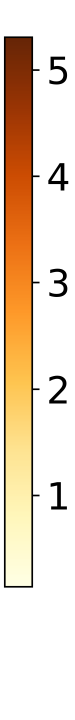}}
	\caption{Heatmap showing, for each stiffness target, the quantity $\frac{1}{d_\text{closest}}$, with $d_\text{closest}$ being the distance from the closest of the collected 18285 simulations. Each sub-figure is related to a pair of stiffness values: (a) $k_\xi, k_\eta$; (b) $k_\xi, k_\zeta$; (c) $k_\eta, k_\zeta$.}
	\label{fig:heatmap}
\end{figure}

We collected the results of all simulations carried out plus additional runs aiming at target configurations that were never observed, so as to explore in depth the parameters space and confirm that missing target configurations are actually unreachable. With an overall collection of 18285 simulations results, we were finally able to plot (Figure \ref{fig:heatmap}) which parts of the three-dimensional target space are ultimately reachable. In particular, some regions where at least one of the three stiffness values is extreme are not physically obtainable; moreover, some stiffness values are obtainable, but with very precise configurations that are on the edge of physical feasibility (e.g., $(k_\xi, k_\eta, k_\zeta) = (2522.46, 139.86, 15.58)$).

\section{Conclusion} 
\label{sec:conclusion}
In this work, we presented an optimization-based approach to identify the geometric properties of the WaveJoint that yield the desired stiffness. By modeling the task as a black-box global optimization problem and leveraging a surrogate-based strategy, we demonstrated the method's effectiveness in minimizing the number of required measurements while handling geometric constraints.  

We show that the proposed algorithm identifies optimal parameters for the WaveJoint in both simulated and real-world scenarios, even with no initial data. Furthermore, as the number of measurements increases, the method transitions from iterative optimization to providing a direct mapping, enhancing its applicability in different experimental contexts.  

This approach provides a flexible and efficient tool for exploring complex geometric parameter spaces, with potential applications extending beyond the WaveJoint to other soft robotic systems. Future work will focus on extending this methodology to other types of joints and exploring its integration into broader design frameworks for robotic systems.


\section*{Competing Interest}
The authors have no competing interests to declare that are relevant to the content of this article.

\section*{Acknowledgements} 
We acknowledge the support of the European Union by the Next Generation EU project ECS00000017 ``Ecosistema dell’Innovazione'' Tuscany Health Ecosystem (THE, PNRR: Spoke 9 - Robotics and Automation for Health), and by the Horizon Europe project ``HARIA - Human-Robot Sensorimotor Augmentation - Wearable Sensorimotor Interfaces and Super- numerary Robotic Limbs for Humans with Upper-limb Disabilities'' (GA No. 101070292), and Italian Ministry of Research, under the complementary actions to the NRRP ``Fit4MedRob - Fit for Medical Robotics'' Grant (PNC0000007, CUP B53C22006960001).



\bibliographystyle{abbrv}

\begin{thebibliography}{10}
	
	\bibitem{deimel2013compliant}
	R.~Deimel and O.~Brock, ``A compliant hand based on a novel pneumatic
	actuator,'' in {\em Robotics and Automation (ICRA), 2013 IEEE International
		Conference on}, pp.~2047--2053, IEEE, 2013.
	
	\bibitem{catalano2014adaptive}
	M.~G. Catalano, G.~Grioli, E.~Farnioli, A.~Serio, C.~Piazza, and A.~Bicchi,
	``Adaptive synergies for the design and control of the pisa/iit softhand,''
	{\em The International Journal of Robotics Research}, vol.~33, no.~5,
	pp.~768--782, 2014.
	
	\bibitem{feng2018soft}
	N.~Feng, Q.~Shi, H.~Wang, J.~Gong, C.~Liu, and Z.~Lu, ``A soft robotic hand:
	design, analysis, semg control, and experiment.,'' {\em International Journal
		of Advanced Manufacturing Technology}, vol.~97, 2018.
	
	\bibitem{shintake2018soft}
	J.~Shintake, V.~Cacucciolo, D.~Floreano, and H.~Shea, ``Soft robotic
	grippers,'' {\em Advanced Materials}, vol.~30, no.~29, p.~1707035, 2018.
	
	\bibitem{Bicchi2000TR0}
	A.~Bicchi, ``Hands for dextrous manipulation and robust grasping: a difficult
	road towards simplicity,'' {\em IEEE Trans. on Robotics and Automation},
	vol.~16, pp.~652--662, December 2000.
	
	\bibitem{Prattichizzo08}
	D.~Prattichizzo and J.~Trinkle, ``Grasping,'' in {\em Handbook on Robotics}
	(B.~Siciliano and O.~Kathib, eds.), pp.~671--700, Springer, 2008.
	
	\bibitem{PoMaSaBiMaPr_IJRR2020}
	M.~Pozzi, S.~Marullo, G.~Salvietti, J.~Bimbo, M.~Malvezzi, and D.~Prattichizzo,
	``Hand closure model for planning top grasps with soft robotic hands,'' {\em
		The International Journal of Robotics Research}, vol.~39, no.~14,
	pp.~1706--1723, 2020.
	
	\bibitem{hussain2020design}
	I.~Hussain, O.~Al-Ketan, F.~Renda, M.~Malvezzi, D.~Prattichizzo,
	L.~Seneviratne, R.~K. Abu Al-Rub, and D.~Gan, ``Design and prototyping
	soft--rigid tendon-driven modular grippers using interpenetrating phase
	composites materials,'' {\em The International Journal of Robotics Research},
	vol.~39, no.~14, pp.~1635--1646, 2020.
	
	\bibitem{jones2021bubble}
	T.~J. Jones, E.~Jambon-Puillet, J.~Marthelot, and P.-T. Brun, ``Bubble casting
	soft robotics,'' {\em Nature}, vol.~599, no.~7884, pp.~229--233, 2021.
	
	\bibitem{della2018toward}
	C.~Della~Santina, C.~Piazza, G.~Grioli, M.~G. Catalano, and A.~Bicchi, ``Toward
	dexterous manipulation with augmented adaptive synergies: The pisa/iit
	softhand 2,'' {\em IEEE Transactions on Robotics}, vol.~34, no.~5,
	pp.~1141--1156, 2018.
	
	\bibitem{PoSaBiMaPr-ral2018}
	M.~Pozzi, G.~Salvietti, J.~Bimbo, M.~Malvezzi, and D.~Prattichizzo, ``The
	closure signature: A functional approach to model underactuated compliant
	robotic hands,'' {\em IEEE Robotics and Automation Letters}, vol.~3,
	pp.~2206--2213, July 2018.
	
	\bibitem{odhner2014compliant}
	L.~U. Odhner, L.~P. Jentoft, M.~R. Claffee, N.~Corson, Y.~Tenzer, R.~R. Ma,
	M.~Buehler, R.~Kohout, R.~D. Howe, and A.~M. Dollar, ``A compliant,
	underactuated hand for robust manipulation,'' {\em The International Journal
		of Robotics Research}, vol.~33, no.~5, pp.~736--752, 2014.
	
	\bibitem{dollar2010highly}
	A.~M. Dollar and R.~D. Howe, ``The highly adaptive sdm hand: Design and
	performance evaluation,'' {\em The international journal of robotics
		research}, vol.~29, no.~5, pp.~585--597, 2010.
	
	\bibitem{Salvietti-Robosoft2020}
	A.~Gafer, D.~Heymans, D.~Prattichizzo, and G.~Salvietti, ``The quad-spatula
	gripper: A novel soft-rigid gripper for food handling,'' in {\em Proc. IEEE
		Int. Conf. on Soft Robotics (RoboSoft)}, (New Haeven, USA), pp.~39--45, 2020.
	
	\bibitem{malvezzi2019design}
	M.~Malvezzi, Z.~Iqbal, M.~C. Valigi, M.~Pozzi, D.~Prattichizzo, and
	G.~Salvietti, ``Design of multiple wearable robotic extra fingers for human
	hand augmentation,'' {\em Robotics}, vol.~8, no.~4, p.~102, 2019.
	
	\bibitem{SaHuMaPr-RAL2017}
	G.~Salvietti, I.~Hussain, M.~Malvezzi, and D.~Prattichizzo, ``Design of the
	passive joints of underactuated modular soft hands for fingertip trajectory
	tracking,'' {\em {IEEE Robotics and Automation Letters}}, vol.~2, no.~4,
	pp.~2008--2015, 2017.
	
	\bibitem{HuIqMaPrSa_HFR19}
	I.~Hussain, Z.~Iqbal, M.~Malvezzi, D.~Prattichizzo, and G.~Salvietti, ``How to
	3d-print compliant joints with a selected stiffness for cooperative
	underactuated soft grippers,'' in {\em Human-Friendly Robotics 2019. HFR
		2019} (F.~Ferraguti, V.~Villani, L.~Sabattini, and M.~Bonfé, eds.), vol.~12,
	pp.~139--153, Springer, Cham, 2020.
	
	\bibitem{xu2018bend}
	H.~Xu, E.~Knoop, S.~Coros, and M.~B{\"a}cher, ``Bend-it: design and fabrication
	of kinetic wire characters,'' {\em ACM Transactions on Graphics (TOG)},
	vol.~37, no.~6, pp.~1--15, 2018.
	
	\bibitem{megaro2017computational}
	V.~Megaro, J.~Zehnder, M.~B{\"a}cher, S.~Coros, M.~Gross, and B.~Thomaszewski,
	``A computational design tool for compliant mechanisms,'' {\em ACM
		Transactions on Graphics (TOG)}, vol.~36, no.~4, pp.~1--12, 2017.
	
	\bibitem{mutlu20173d}
	R.~Mutlu, C.~Tawk, G.~Alici, and E.~Sariyildiz, ``A 3d printed monolithic soft
	gripper with adjustable stiffness,'' in {\em IECON 2017-43rd Annual
		Conference of the IEEE Industrial Electronics Society}, pp.~6235--6240, IEEE,
	2017.
	
	\bibitem{8868668}
	C.~Tawk, Y.~Gao, R.~Mutlu, and G.~Alici, ``Fully 3d printed monolithic soft
	gripper with high conformal grasping capability,'' in {\em 2019 IEEE/ASME
		International Conference on Advanced Intelligent Mechatronics (AIM)},
	pp.~1139--1144, 2019.
	
	\bibitem{mutlu2015effect}
	R.~Mutlu, G.~Alici, M.~in~het Panhuis, and G.~Spinks, ``Effect of flexure hinge
	type on a 3d printed fully compliant prosthetic finger,'' in {\em 2015 IEEE
		International Conference on Advanced Intelligent Mechatronics (AIM)},
	pp.~790--795, IEEE, 2015.
	
	\bibitem{liu2018soft}
	C.-H. Liu, C.-H. Chiu, T.-L. Chen, T.-Y. Pai, Y.~Chen, and M.-C. Hsu, ``A soft
	robotic gripper module with 3d printed compliant fingers for grasping
	fruits,'' in {\em 2018 IEEE/ASME International Conference on Advanced
		Intelligent Mechatronics (AIM)}, pp.~736--741, IEEE, 2018.
	
	\bibitem{dragusanu2022wavejoints}
	M.~Dragusanu, G.~M. Achilli, M.~C. Valigi, D.~Prattichizzo, M.~Malvezzi, and
	G.~Salvietti, ``The wavejoints: A novel methodology to design soft-rigid
	grippers made by monolithic 3d printed fingers with adjustable joint
	stiffness,'' in {\em 2022 International Conference on Robotics and Automation
		(ICRA)}, pp.~6173--6179, IEEE, 2022.
	
	\bibitem{10160353}
	M.~Dragusanu, D.~Troisi, D.~Prattichizzo, and M.~Malvezzi, ``Compliant finger
	joint with controlled variable stiffness based on twisted strings
	actuation,'' in {\em 2023 IEEE International Conference on Robotics and
		Automation (ICRA)}, pp.~7378--7384, 2023.
	
	\bibitem{gutmann2001radial}
	H.-M. Gutmann, ``A radial basis function method for global optimization,'' {\em
		Journal of Global Optimization}, vol.~19, no.~3, pp.~201--227, 2001.
	
	\bibitem{costa2018rbfopt}
	A.~Costa and G.~Nannicini, ``Rbfopt: an open-source library for black-box
	optimization with costly function evaluations,'' {\em Mathematical
		Programming Computation}, vol.~10, pp.~597--629, 2018.
	
	\bibitem{cassioli2013global}
	A.~Cassioli and F.~Schoen, ``Global optimization of expensive black box
	problems with a known lower bound,'' {\em Journal of Global Optimization},
	vol.~57, pp.~177--190, 2013.
	
	\bibitem{jones1998efficient}
	D.~R. Jones, M.~Schonlau, and W.~J. Welch, ``Efficient global optimization of
	expensive black-box functions,'' {\em Journal of Global optimization},
	vol.~13, pp.~455--492, 1998.
	
	\bibitem{jones2001taxonomy}
	D.~R. Jones, ``A taxonomy of global optimization methods based on response
	surfaces,'' {\em Journal of Global Optimization}, vol.~21, no.~4,
	pp.~345--383, 2001.
	
\end{thebibliography}

\renewcommand{\theequation}{A.\arabic{equation}}
\setcounter{equation}{0}

\renewcommand{\thetable}{B.\Roman{table}}
\setcounter{table}{0}
\renewcommand{\theHtable}{Supplement\thetable}

\renewcommand{\thealgocf}{A.\arabic{algocf}}
\setcounter{algocf}{0}

{\appendices
	\section{Radial Basis Function methods for global optimization}\label{app::A}
	A \textit{radial function} $s(x)$ \cite{gutmann2001radial} is defined as
	\begin{equation}
		s(x) := \sum_{i=1}^{h} \lambda_i \sigma(\|x-\bar{x}_i\|),
		\label{eq:surrogate_interpolant}
	\end{equation}
	where $\lambda_1, \dots, \lambda_h \in\mathbb{R}$ are scalar coefficients, $\bar{x}_1, \dots, \bar{x}_h \in\Omega$ are the centers of the radial bases, $\Omega$ is the feasible set, $\|\cdot\|$ is the Euclidean norm and $\sigma : \mathbb{R}^+ \to \mathbb{R}$ is a univariate function usually referred to as \textit{radial basis function} (RBF) - the Gaussian RBF $\sigma(r)=\exp(-\gamma r^2)$ is arguably the most popular one.
	A radial function is thus expressed as a linear combination of RBFs,
	each term being symmetric around its center. A very
	common choice in interpolation tasks is to choose the set of centers
	to coincide with the interpolation points. 
	
	Let $\mathcal{D} = \{(x_i, y_i) \in \Omega \times \mathbb{R} \;|\; i = 1,\dots,N\}$ be a set of $N$ observations for which $y_i = f(x_i)$, $i = 1, \dots, N$. Denoting by
	$\Xi$ the (symmetric) matrix whose $(i, j)$-th entry is $\sigma(\| x_i - x_j \|)$, $\forall i, j = 1, \dots, N$, the problem of finding the interpolant \eqref{eq:surrogate_interpolant} can be formulated as a linear system: \begin{equation}
		\Xi \lambda = y.
		\label{eq:rbf_linear_system}
	\end{equation}
	A uniquely defined radial basis interpolant exists if and only if the linear system has a unique solution in $\lambda$, i.e., if and only if matrix $\Xi$ is invertible. Depending on the choice of the radial basis function, this condition might not be satisfied. 
	The common way used to guarantee the existence of a solution of the system is to add a low degree polynomial to the interpolant.
	
	Approaches based on RBFs can be used to tackle problems of the general form
	\begin{equation}
		\label{eq:opt_prob_gen}
		\begin{aligned}
			\min_{x}\;& f(x) = \|\phi(x)-T\|^2\\\text{s.t. }&x\in[\ell,\mu],\; x\in \mathbb{R}^n\times\mathbb{Z}^m,
		\end{aligned}
	\end{equation}
	where $\ell,\mu\in\mathbb{R}^{n+m}$ and we denote the overall feasible set by $\Omega = [\ell,\mu]\cap \mathbb{R}^n\times \mathbb{Z}^m$.
	Algorithm \ref{alg:general_scheme} shows the general procedure adopted in global optimization based on RBFs. Usually the algorithm ends when a certain number of function evaluations has been performed or a time limit is reached, returning the best point, i.e., the observation with the lowest function value found so far.
	
	\begin{algorithm}[H]
		\caption{Global Optimization Algorithm Based on RBF surrogate}
		\label{alg:general_scheme}
		\begin{algorithmic}[1]
			\renewcommand{\algorithmicrequire}{\textbf{Input:}}
			\REQUIRE Lower and upper bounds $\ell\in\mathbb{R}^{n+m}$, $\mu\in\mathbb{R}^{n+m}$, $N$ number of initial points
			\STATE Choose initial points $x_1, \dots, x_N \in\Omega$ and evaluate $f$  obtaining $y_1, \dots, y_N$
			\STATE Let $\mathcal{D}_x = \{x_i \mid i = 1,\dots,N\}$ and $\mathcal{D}_y = \{y_i \mid i = 1,\dots,N\}$
			\STATE Set $k \leftarrow N$
			\WHILE {\textit{stopping criterion not satisfied}}
			\STATE Compute RBF surrogate $s_k(x)$ on $\mathcal{D}_x$ and $\mathcal{D}_y$ by solving system \eqref{eq:rbf_linear_system}
			\STATE \label{step:min_surr} Minimize $s_k(x)$  and get
			$x_{k+1}$
			\STATE Evaluate $f(x_{k+1})$ obtaining $y_{k+1}$
			\STATE Update $\mathcal{D}_x  = \mathcal{D}_x  \; \cup \; \{x_{k+1}\}$ and $\mathcal{D}_y  = \mathcal{D}_y  \; \cup \; \{y_{k+1}\}$
			\STATE Set $k \leftarrow k + 1$
			\ENDWHILE
			\STATE Let $x^* = \argmin_{x\in\mathcal{D}_x} f(x)$
			\RETURN $x^*, f(x^*)$
		\end{algorithmic}
	\end{algorithm}
	
	In fact, instead of minimizing $s_k(x)$, more advanced variants of Algorithm \ref{alg:general_scheme} employ merit (or acquisition) functions to be minimized for obtaining $x_{k+1}$. An acquisition function $m_k(x) = s_k(x)+u_k(x)$ shall calibrate the trade-off between \textit{exploration} of the feasible region (measured by an uncertainty function $u_k(x)$) and the \textit{exploitation} of the surrogate model $s_k(x)$ of $f(x)$.

	\section{Extended Numerical Results}\label{app::B}
	
	In this section, we report numerical results which did not find space in the main body of the manuscript. In particular: all the experiments for Table \ref{tab::preliminary} of the manuscript are reported in Tables \ref{tab:f_before}-\ref{tab:f_after}; Table \ref{tab:incremental_app} contains the results of the entire experimentation performed for Table \ref{tab:incremental} of the paper.
	
	\begin{table*}[!h]
		\centering%
		\renewcommand{\arraystretch}{1.2}%
		\caption{Results of WaveJoints design by RBFOpt (no data at initialization), according to FEM simulations.}
		\label{tab:f_before}
		\scriptsize
		\begin{tabular}{|c||ccccc||ccc||c||c|}%
			\hline%
			\multirow{2}{*}{Type}&\multicolumn{5}{c||}{Configuration}&\multicolumn{3}{c||}{Stiffness}&\multirow{2}{*}{$\mathcal{R}$}&\multirow{2}{*}{Time (h)}\\%
			\cline{2%
				-%
				9}%
			&$l_t$&$n_r$&$h_t$&$t_h$&$\alpha$&$k_\xi$&$k_\eta$&$k_\zeta$&&\\%
			\hline%
			\hline%
			Target&15.0&3.0&8.0&0.5&0.0&990.2987435&35.2796&5.0932&\multirow{2}{*}{5e{-}05}&\multirow{2}{*}{2.59}\\%
			\cline{1%
				-%
				9}%
			Final&15.03&3.0&8.03&0.5&0.0&994.7744533&35.1521&5.1109&&\\%
			\hline%
			\hline%
			Target&15.0&4.0&12.0&0.5&0.0&1368.4751095&25.5473&3.8731&\multirow{2}{*}{0.00192}&\multirow{2}{*}{2.63}\\%
			\cline{1%
				-%
				9}%
			Final&18.91&5.0&10.91&0.54&5.0&1335.141762&26.4364&3.8322&&\\%
			\hline%
			\hline%
			Target&20.0&3.0&14.0&0.5&0.0&741.2807955&6.0003&1.9733&\multirow{2}{*}{3e{-}05}&\multirow{2}{*}{2.63}\\%
			\cline{1%
				-%
				9}%
			Final&27.95&4.0&11.66&0.59&11.0&744.9876164&6.0016&1.9776&&\\%
			\hline%
			\hline%
			Target&20.0&5.0&10.0&0.5&3.0&933.9547194&16.6476&2.4329&\multirow{2}{*}{0.00111}&\multirow{2}{*}{2.56}\\%
			\cline{1%
				-%
				9}%
			Final&27.55&6.0&9.25&0.6&5.0&962.0831036&16.4722&2.4563&&\\%
			\hline%
			\hline%
			Target&20.0&4.0&10.0&0.5&12.0&794.4973146&8.5653&2.4449&\multirow{2}{*}{0.00251}&\multirow{2}{*}{2.5}\\%
			\cline{1%
				-%
				9}%
			Final&22.73&4.0&9.62&0.54&12.0&761.6266629&8.6578&2.5089&&\\%
			\hline%
			\hline%
			Target&20.0&4.0&10.0&0.5&21.0&769.2113467&5.9468&2.5417&\multirow{2}{*}{0.00023}&\multirow{2}{*}{2.6}\\%
			\cline{1%
				-%
				9}%
			Final&20.24&4.0&10.16&0.51&21.0&779.0096133&5.9011&2.5487&&\\%
			\hline%
			\hline%
			Target&20.0&3.0&10.0&0.5&21.0&635.4721141&6.2179&2.7956&\multirow{2}{*}{0.0}&\multirow{2}{*}{1.1}\\%
			\cline{1%
				-%
				9}%
			Final&27.42&3.0&10.64&0.6&20.0&635.9104309&6.2123&2.7925&&\\%
			\hline%
		\end{tabular}%
	\end{table*}
	
	\begin{table*}[!h]
		\centering
		\renewcommand{\arraystretch}{1.2}%
		\caption{Results of WaveJoints design by RBFOpt (3043 data points at initialization), according to FEM simulations.}
		\label{tab:f_after}
		\scriptsize
		\begin{tabular}{|c||ccccc||ccc||c||c|}%
			\hline%
			\multirow{2}{*}{Type}&\multicolumn{5}{c||}{Configuration}&\multicolumn{3}{c||}{Stiffness}&\multirow{2}{*}{$\mathcal{R}$}&\multirow{2}{*}{Time (h)}\\%
			\cline{2%
				-%
				9}%
			&$l_t$&$n_r$&$h_t$&$t_h$&$\alpha$&$k_\xi$&$k_\eta$&$k_\zeta$&&\\%
			\hline%
			\hline%
			Target&15.0&3.0&12.0&0.5&0.0&1153.1955404&15.8598&3.7215&\multirow{2}{*}{0.0012}&\multirow{2}{*}{5.8}\\%
			\cline{1%
				-%
				9}%
			Final&15.26&3.0&11.57&0.5&1.0&1121.7993543&16.1922&3.7361&&\\%
			\hline%
			\hline%
			Target&20.0&4.0&10.0&0.5&0.0&803.3871624&12.4934&2.3771&\multirow{2}{*}{0.00087}&\multirow{2}{*}{4.89}\\%
			\cline{1%
				-%
				9}%
			Final&27.05&5.0&9.08&0.58&5.0&818.7625301&12.6985&2.3406&&\\%
			\hline%
			\hline%
			Target&25.0&3.0&10.0&0.5&0.0&448.0967673&7.1477&1.8049&\multirow{2}{*}{0.00045}&\multirow{2}{*}{3.5}\\%
			\cline{1%
				-%
				9}%
			Final&28.67&4.0&8.0&0.53&10.0&456.494794&7.0798&1.8093&&\\%
			\hline%
			\hline%
			Target&20.0&5.0&10.0&0.5&6.0&932.272608&14.2617&2.4533&\multirow{2}{*}{3e{-}05}&\multirow{2}{*}{4.79}\\%
			\cline{1%
				-%
				9}%
			Final&27.86&6.0&8.99&0.6&7.0&936.2972516&14.2246&2.4512&&\\%
			\hline%
			\hline%
			Target&20.0&4.0&10.0&0.5&15.0&785.8311582&7.5199&2.4704&\multirow{2}{*}{0.00059}&\multirow{2}{*}{3.67}\\%
			\cline{1%
				-%
				9}%
			Final&22.91&5.0&8.64&0.52&16.0&776.9848613&7.5952&2.4232&&\\%
			\hline%
			\hline%
			Target&20.0&4.0&10.0&0.5&24.0&760.010985&5.3672&2.5943&\multirow{2}{*}{0.00037}&\multirow{2}{*}{5.0}\\%
			\cline{1%
				-%
				9}%
			Final&24.39&3.0&13.57&0.59&20.0&771.3502553&5.3016&2.5941&&\\%
			\hline%
			\hline%
			Target&20.0&3.0&10.0&0.5&24.0&630.0396666&5.7302&2.8581&\multirow{2}{*}{1e{-}05}&\multirow{2}{*}{4.61}\\%
			\cline{1%
				-%
				9}%
			Final&20.01&3.0&9.98&0.5&24.0&629.2406172&5.724&2.866&&\\%
			\hline%
		\end{tabular}%
	\end{table*}
	
	\begin{table*}[!h]
		\centering
		\renewcommand{\arraystretch}{1.2}%
		\caption{Evolution of RBFOpt process at ``milestone'' iterations where a module is physically 3D-printed. The target stiffness values are $(k_\xi, k_\eta, k_\zeta) = (1238.27, 50.84, 5)$.}
		\label{tab:incremental_app}
		\scriptsize
		\begin{tabular}{|c|c||c|ccc|c|}%
			\hline%
			\multirow{2}{*}{N$^{\circ}$ Simul.}&\multirow{2}{*}{N$^{\circ}$ Real}&\multirow{2}{*}{Type}&\multicolumn{3}{c|}{Stiffness}&\multirow{2}{*}{$\mathcal{R}$}\\%
			\cline{4%
				-%
				6}%
			&&&$k_\xi$&$k_\eta$&$k_\zeta$&\\%
			\hline%
			\hline%
			\multirow{2}{*}{25}&\multirow{2}{*}{1}&Simul.&1652.89&31.07&4.63&0.268979\\%
			\cline{3%
				-%
				7}%
			&&Real&2683.5&47.46&11.25&2.922401\\%
			\hline%
			\hline%
			\multirow{2}{*}{50}&\multirow{2}{*}{2}&Simul.&1590.97&25.5&4.3&0.349188\\%
			\cline{3%
				-%
				7}%
			&&Real&1451.7&23.89&11.29&1.887757\\%
			\hline%
			\hline%
			\multirow{2}{*}{75}&\multirow{2}{*}{3}&Simul.&1641.87&30.18&5.16&0.272236\\%
			\cline{3%
				-%
				7}%
			&&Real&1383.7&30.43&6.21&0.232495\\%
			\hline%
			\hline%
			\multirow{2}{*}{100}&\multirow{2}{*}{4}&Simul.&1641.87&30.18&5.16&0.272236\\%
			\cline{3%
				-%
				7}%
			&&Real&1383.7&30.43&6.21&0.232495\\%
			\hline%
			\hline%
			\multirow{2}{*}{125}&\multirow{2}{*}{5}&Simul.&1517.04&29.58&4.96&0.225565\\%
			\cline{3%
				-%
				7}%
			&&Real&1581.4&28.57&5.71&0.288711\\%
			\hline%
			\hline%
			\multirow{2}{*}{150}&\multirow{2}{*}{6}&Simul.&1428.97&28.59&4.51&0.225254\\%
			\cline{3%
				-%
				7}%
			&&Real&1369.8&27.45&5.32&0.226888\\%
			\hline%
			\hline%
			\multirow{2}{*}{175}&\multirow{2}{*}{7}&Simul.&1507.89&42.84&5.9&0.104348\\%
			\cline{3%
				-%
				7}%
			&&Real&1272.0&28.57&5.5&0.202557\\%
			\hline%
			\hline%
			\multirow{2}{*}{200}&\multirow{2}{*}{8}&Simul.&1353.96&37.0&5.56&0.09524\\%
			\cline{3%
				-%
				7}%
			&&Real&1118.4&32.85&5.05&0.134661\\%
			\hline%
		\end{tabular}%
	\end{table*}
}

\end{document}